\documentclass{tlp}

\usepackage{aopmath}

\usepackage{verbatim}
\usepackage{amssymb}
\usepackage{tikz}
\usetikzlibrary{arrows}

\usepackage{subfigure}

\usepackage{stmaryrd}

\newtheorem{definition}{Definition} 
\newtheorem{example}{Example} 
\newtheorem{remark}{Remark}
\newtheorem{problem}{Problem}

\newcommand{\revision}{\ensuremath{\star}}
\newcommand{\allinterpretations}{\ensuremath{\Omega}}
\newcommand\modeles[1]{\ensuremath{mod(#1)}}
\newcommand{\Y}{\ensuremath{(Y, Y)}}

\newcommand{\Yppsingle}{\ensuremath{Y^{(2)}}}

\newcommand{\XY}{\ensuremath{(X, Y)}}

\newcommand{\Yetoile}{\ensuremath{(Y_*, Y_*)}}

\newcommand\formwithoutbraces[1]{\ensuremath{\mathsf{lp}(#1)}}
\newcommand\form[1]{\ensuremath{\mathsf{lp}(\{#1\})}}
\newcommand\formY{\ensuremath{\form{Y}}}

\newcommand\formYYp{\ensuremath{\form{Y, Y'}}}
\newcommand\formYYpp{\ensuremath{\form{Y, Y^{(2)}}}}
\newcommand\formYpYpp{\ensuremath{\form{Y', Y^{(2)}}}}
\newcommand\formYYpYpp{\ensuremath{\form{Y, Y', Y^{(2)}}}}
\newcommand\formXYY{\ensuremath{\form{(X, Y), (Y, Y)}}}
\renewcommand{\S}{\ensuremath{\mathsf{S}}}
\newcommand\GLPops[1]{\ensuremath{GLP(#1)}}
\newcommand\lattice[1]{\preceq_{#1}}

\newcommand\formsymb{\ensuremath{\alpha}}
\newcommand\formGLPb[1]{\ensuremath{\formsymb^2_{#1}}}
\newcommand\formGLPa[1]{\ensuremath{\formsymb^1_{#1}}}

\newcommand\trichea{\vspace*{-.1cm}}

\renewcommand\P{\ensuremath{\mathcal{P}}}
\newcommand\Q{\ensuremath{\mathcal{Q}}}
\newcommand\R{\ensuremath{\mathcal{R}}}

\newcommand\SE{\ensuremath{\mathcal{SE}}}

\newcommand\Rsharp{\ensuremath{\mathcal{R}_\#}}

\newcommand{\tqed}{\rule{1.5mm}{1.5mm}}

\newcommand\nico[1]{#1}

\begin{document}
\bibliographystyle{acmtrans}

\long\def\comment#1{}

\title{Characterization of Logic Program Revision as an Extension of Propositional Revision\footnote{This is a revised and full version (including proofs of propositions) of \cite{DBLP:conf/lpnmr/SchwindI13}.}}
\shorttitle{Revising Logic Programs: Characterization Results}

\author[N. Schwind \and K. Inoue]{
	Nicolas Schwind\\
	Transdisciplinary Research Integration Center\\
	National Institute of Informatics\\
	2-1-2 Hitotsubashi, Chiyoda-ku\\
	101-8430 Tokyo, Japan\\
	E-mail: schwind@nii.ac.jp
	\and Katsumi Inoue \\
	National Institute of Informatics\\
	2-1-2 Hitotsubashi, Chiyoda-ku\\
	101-8430 Tokyo, Japan\\
	E-mail: inoue@nii.ac.jp
}

\submitted{8 June 2014}
\revised{12 May 2015}
\accepted{22 June 2015}

\pagerange{\pageref{firstpage}--\pageref{lastpage}}
\setcounter{page}{1}

\hyphenation{
mo-de-ling
ge-ne-ra-li-zed
ta-king
a-xio-ma-ti-cal-ly
o-pe-ra-tors
o-pe-ra-tor
or-de-ring
ac-cor-ding
cor-res-ponds
se-cond
dif-fe-rent
ve-ri-fied
res-pec-ti-ve-ly
re-pre-sen-ta-tion
e-xists
e-qui-va-len-ce
}

\maketitle

\label{firstpage}

\noindent\textbf{\textit{Note: This article has been accepted for publication and will appear in Theory and Practice of Logic Programming.}}\\


\begin{abstract}
We address the problem of belief revision of logic programs, i.e., how to incorporate to a logic program $\P$ a new
logic program $\Q$. Based on the structure of SE interpretations,
Delgrande \textit{et al.\/} \citeyear{DBLP:conf/kr/DelgrandeSTW08,Delgrande/ACM} adapted the well-known AGM framework
\citeyear{AGM85} to logic program (LP) revision.
They identified the
rational behavior of LP revision and introduced some specific operators.
In this paper, a constructive characterization of all rational LP revision operators
is given in terms of orderings over propositional interpretations with some further conditions specific to SE interpretations.
It provides an intuitive, complete procedure for the construction of all rational LP revision operators and makes easier the comprehension of their semantic and
computational properties.
We give a particular consideration to logic programs of very general form, i.e., the \emph{generalized} logic programs (GLPs).
We show that every rational GLP revision operator is derived from a propositional revision operator satisfying the original AGM postulates.
Interestingly, the further conditions specific to GLP revision are independent from the propositional revision operator on which a GLP revision operator
is based.
Taking advantage of our characterization result, we embed the GLP revision operators into structures of Boolean lattices, that allow us to bring to
light some potential weaknesses in the adapted AGM postulates. To illustrate our claim, we introduce and characterize axiomatically two specific classes
of (rational) GLP revision operators which arguably have a drastic behavior.
We additionally consider two more restricted forms of logic programs, i.e., the \emph{disjunctive} logic programs (DLPs) and the \emph{normal}
logic programs (NLPs) and adapt our characterization result to DLP and NLP revision operators.

	
\end{abstract}


\section{Introduction}
\label{sec:introduction}

Logic programs (LPs) under the answer set semantics are well-suited
for modeling
problems which involve common sense
reasoning (e.g., biological networks, diagnosis, planning, etc.)
Due to the dynamic nature of
our environment, beliefs represented through an LP
$\P$ are subject to change, i.e., because one wants to
incorporate to it a new LP $\Q$.
Since there is no unique, consensual procedure to revise a set of beliefs,
Alchourr\'on, G\"ardenfors and Makinson \citeyear{AGM85} introduced a set of desirable principles
w.r.t. belief change called \textit{AGM postulates}.
Katsuno and Mendelzon
\citeyear{KatsunoMendelzon92} adapted
these principles to the case of propositional logic,
distinguished two kind of change operations, i.e.,
\textit{revision} and \textit{update} \cite{KMupdate},
and characterized axiomatically each one of these change operations by
a set of so-called \textit{KM postulates}.
Revision consists in incorporating a new information
into a database that represents a static world, i.e.,
new and old beliefs describe the same situation but new ones
are more reliable.
In the case of update,
the underlying world
evolves with respect to the occurence of some events
i.e., new and old beliefs describe
two different states of the world.

Our interests focus here on the problem of
revision
of logic programs.
Most of works dealing with belief change in logic programming are concerned with
rule-based update \cite{DBLP:conf/ijcai/ZhangF97,DBLP:journals/jlp/AlferesLPPP00,DBLP:journals/tplp/EiterFST02,DBLP:journals/tplp/SakamaI03,journals/tocl/Zhang06,DBLP:conf/lpnmr/DelgrandeST07}, and they do not lie into the AGM framework, particularly due
to their syntactic essence.

Indeed, given the
nonmonotonic nature of LPs
the AGM/KM postulates can not be
directly applied to logic programs \cite{DBLP:journals/tplp/EiterFST02}.
However,
the notion of \textit{SE models} introduced by Turner \citeyear{DBLP:journals/tplp/Turner03}
provided a monotonic semantical characterization of LPs, which is more expressive than the answer set semantics.
Initially, SE models were used to characterize the strong equivalence between logic programs \cite{Lifschitz:2001:SEL:383779.383783}:
\nico{
precisely, two LPs have the same set of SE models if and only if they are strongly equivalent, that is to say, they admit the same answer
sets, and will still do even after adding any arbitrary set of rules to them.
}

Based on these structures, Delgrande \textit{et al.\/} \citeyear{DBLP:conf/kr/DelgrandeSTW08,Delgrande/ACM}
adapted the AGM/KM postulates in the context of logic programming.
They focused on the revision of logic programs, i.e., they proposed several revision
operators and investigated their properties
w.r.t. the adapted postulates.
\nico{
Slota and Leite \citeyear{Slota:2010:SUO:1860967.1861155,DBLP:journals/corr/SlotaL13} exploited the same idea for
\textit{update} of logic programs by adapting the KM postulates in a similar way.
These semantical-based belief change operations (revision and update) changed the focus from the dynamic evolution of a
syntactic, rule-based representation of beliefs previously proposed in the literature to the evolution
of its semantic content; these works
covered a serious drawback in the field of belief revision in logic programming.
In the context of update, Slota and Leite also
proposed a
\textit{constructive representation}
of such update operators.
Such a result provides a sound and complete model-theoretic construction of the rational LP update operators,
i.e., a ``generic recipe'' to construct all operators that fully satisfy the adaptation of the AGM/KM postulates to logic programs. It is indeed crucial
when defining a logical operator in an axiomatic way to give an intuitive constructive characterization of it
in order to aid the analysis of its semantic and computational properties.
}

In this paper,
we give a particular consideration to the revision of 
\emph{generalized} logic programs (GLPs) \cite{DBLP:journals/jlp/InoueS98}
which are of very general form.
Revising a GLP $\P$
by an other GLP $\Q$ should result in a new GLP
that satisfy the adapted set of AGM postulates.
We provide a characterization of the set of all GLP revision operators
by associating each GLP with a certain structure, called \emph{GLP parted assignment}, which consists of a pair of
assignments that are independent from each other.
Interestingly, the first one, called here \textit{LP faithful assignment}, is similar to the structure of faithful assignment
defined in \cite{KatsunoMendelzon92} and used to characterize the (rational) KM revision operators in the propositional setting; the second one,
called here \textit{well-defined assignment}, can be defined independently from the first one. As a consequence, the benefit of our approach is that:
\begin{itemize}
\item[(i)] every rational LP revision operator $\star$ can be derived from a propositional revision operator $\circ$
satisfying the KM postulates, with some additional conditions that are independent from $\circ$;
\item[(ii)] there is a one-to-one correspondence between the set of rational LP revision operators and the set of all pairs of such assignments.
\end{itemize}

\nico{
Our characterization makes
the refined
analysis of LP revision operators easier.
Indeed,
we can embed the GLP revision operators into structures of Boolean lattices, that allows us to bring out
some potential weaknesses in the original postulates and pave the way for the discrimination of some rational GLP revision operators.
}


The next section introduces some preliminaries about belief revision
in propositional logic. We provide in Section \ref{sec:belief revision logic programming} some necessary background on
generalized logic programs, and we also introduce
the notion on logic program revision, an axiomatic characterization of generalized logic program revision operators, and some
preliminary results.
Section \ref{sec:representation} provides our main result, i.e.,
a constructive characterization of the axiomatic description of
the GLP revision operators.
We formally compare our characterization result with another recent one proposed in \cite{Delgrande/LPNMR};
the benefit of our approach is that our construction is one-to-one, as opposite to Delgrande \textit{et al.\/}'s one.
In Section \ref{sec:lattice} we partition the class of GLP revision operators into subclasses of Boolean lattices,
then we introduce and axiomatically characterize two specific classes
of (rational) GLP revision operators, i.e., the \emph{skeptical} and \emph{brave} GLP revision operators, and lastly
we provide some complexity results which are direct consequences of existing results in the propositional case.
In Section \nolinebreak\ref{sec:disjunctive-normal} we consider the revision of more restricted forms of logic programs, i.e.,
the disjunctive logic programs (DLPs) and normal logic programs (NLPs). We adapt our characterization result to
DLP revision operators and NLP revision operators. Though DLP revision operators and NLP revision operators can also be viewed as
extensions of propositional revision operators, in constrast with GLP revision operators their construction
does not provide us with two independent structures.
We conclude in Section \ref{sec:conclusion}.
The proofs of propositions are provided in an appendix.

This version of the paper is a revised and extended version of a published \linebreak LPNMR'13 paper \cite{DBLP:conf/lpnmr/SchwindI13}. The main extensions include
a comparison of our main characterization result with the one proposed in \cite{Delgrande/LPNMR}, some complexity results,
characterization results for DLP and NLP revision operators and the proofs of propositions.

\section{Belief revision in propositional logic}
\label{sec:belief revision prop logic}
\subsection{Formal preliminaries}

We consider a propositional language ${\cal L}$ defined from a finite set of propositional variables
(also called \emph{atoms}) ${\cal A}$ and the usual connectives.
$\bot$ (resp. $\top$) is the Boolean constant always false (resp. true).
A (classical) \emph{interpretation} over $\mathcal{A}$ is a total function from ${\cal A}$ to $\{0, 1\}$.
To avoid heavy expressions, an interpretation $I$ is also viewed as the subset of atoms from $\mathcal{A}$
that are true in $I$. For instance, if $\mathcal{A} = \{p, q\}$, then the interpretation over $\mathcal{A}$
such that $I(p) = 1$ and $I(q) = 0$ is also represented as the set $\{p\}$.
For the sake of simplicity,
set-notations will be dropped within interpretations (except for the case where the interpretation is the empty set), e.g.,
the interpretation $\{p, q\}$ will be simply denoted $pq$.
The set of all interpretations is denoted $\allinterpretations$.
An interpretation $I$ is a \emph{model} of a formula $\phi \in {\cal L}$, denoted $I \models \phi$, if it makes it true 
in the usual truth functional way. A \emph{consistent} formula is a formula that admits a model.
The set $mod(\phi)$ denotes the set of models of the formula $\phi$, i.e.,  $mod(\phi)=\{I \in \allinterpretations \mid I \models \phi\}$.
Two formulae $\phi, \psi$ are said to be \emph{equivalent}, denoted by $\phi \equiv \psi$ if and only if $mod(\phi) = mod(\psi)$.

\subsection{Propositional revision operators}
We now introduce
some background
on propositional belief revision.
\nico{
We start by introducing a revision operator as a simple function, that considers
two formulae (the original formula and the new one) and that returns the revised formula:
\begin{definition}[Propositional revision operator, equivalence between operators]
A (propositional) revision operator $\circ$ is a mapping associating two formulae $\phi, \psi$ with
a new formula, denoted $\phi \circ \psi$.
Two revision operators $\circ, \circ'$ are said to be \emph{equivalent} (denoted $\circ \equiv \circ'$) when for all formulae
$\phi, \psi$, $\phi \circ \psi \equiv \phi \circ' \psi$. 
\label{def:rev op}
\end{definition}
}

The AGM framework
\cite{AGM85} describes the standard principles for belief revision (e.g., consistency preservation and minimality of change),
which capture changes occuring in a static domain.
Katsuno and Mendelzon \citeyear{KMupdate} equivalently rephrased the AGM postulates as follows:

\begin{definition}[KM revision operator]
A KM revision operator $\circ$ is a propositional revision operator that satisfies the following postulates,
for all formulae $\phi, \phi_1, \phi_2, \psi, \psi_1, \psi_2$:
\newpage
\begin{description}
	\item[\textbf{(R1)}] $\phi \circ \psi \models \psi$;
	\item[\textbf{(R2)}] If $\phi \wedge \psi$ is consistent, then $\phi \circ \psi \equiv \phi \wedge \psi$;
	\item[\textbf{(R3)}] If $\psi$ is consistent, then $\phi \circ \psi$ is consistent;
	\item[\textbf{(R4)}] If $\phi_1 \equiv \phi_2$ and $\psi_1 \equiv \psi_2$, then $\phi_1 \circ \psi_1 \equiv \phi_2 \circ \psi_2$;
	\item[\textbf{(R5)}] $(\phi \circ \psi_1) \wedge \psi_2 \models \phi \circ (\psi_1 \wedge \psi_2)$;
	\item[\textbf{(R6)}] If $(\phi \circ \psi_1) \wedge \psi_2$ is consistent, then $\phi \circ (\psi_1 \wedge \psi_2) \models (\phi \circ \psi_1) \wedge \psi_2$.
\end{description}
\label{def:KM revision operator}
\end{definition}


\nico{
These so-called \emph{KM postulates} capture the desired behavior of a revision operator,
e.g., in terms of consistency preservation and minimality of change.
We now draw the reader's attention to the following important detail.
The KM postulates also tell us that the outcome of a revision operator relies on an arbitrary syntactic distinction:
one can see that a revision operator $\circ$ is a KM revision operator (i.e., it satisfies postulates (R1 - R6)) if and only if
any revision operator equivalent to $\circ$ is also a KM revision operator. In this paper, since we are only interested
in whether an operator satisfies a set of rationality postulates or not, only the semantic contents of the revised base
play a role, that is, relevance is considered only
within the \textit{models} of a revised base rather than on its explicit representation. This is why from now on,
abusing terms we identify a revision operator modulo equivalence, that is, we actually refer to \emph{any} revision operator equivalent to it.
It becomes then harmless to define the resulting revised base in a modelwise fashion, as a set of models implicitely interpreted disjunctively.
As a consequence, given two propositional revision operators $\circ, \circ'$, one can switch between the notations
$\circ \equiv \circ'$ and $\circ = \circ'$ since there is no longer danger of confusion.
}

KM revision operators
can be
represented
in terms of
total preorders over interpretations.
Indeed, each KM revision operator is associated with some faithful assignment \cite{KMupdate}.
For each preorder $\leq$, $\simeq$ denotes the corresponding indifference relation,
and $<$ denotes the corresponding strict ordering;
given a
binary relation
$\leq$ over a set $E$ and any set $F \subseteq E$, the set $\min(F, \leq)$ denotes the subset
of ``minimal'' elements from $F$ w.r.t. $\leq$, i.e.,
$\min(F, \leq) = \{a \in F \mid \forall b \in F, b \leq a \implies a \leq b\}$.

\begin{definition}[Faithful assignment]\label{def:faithful}
A \emph{faithful assignment} is a mapping which associates with every formula $\phi$ a preorder $\leq_\phi$ over
interpretations
such that for all interpretations $I, J$ and all formulae $\phi$, $\phi_1$, $\phi_2$, the following conditions hold:

\begin{description}
\item[(a)] If $I \models \phi$ and $J \models \phi$, then $I \simeq_\phi J$;
\item[(b)] If $I \models \phi$ and $J \not\models \phi$, then $I <_\phi J$;
\item[(c)] If $\phi_1 \equiv \phi_2$, then $\leq_{\phi_1} = \leq_{\phi_2}$.
\end{description}
\end{definition}

\begin{theorem}[Katsuno and Mendelzon 1992]
A revision operator $\circ$ is a KM revision operator if and only if there exists a faithful assignment associating every formula $\phi$ with a total preorder $\leq_\phi$ such that for all formulae $\phi, \psi$, $mod(\phi \circ \psi) = \min(mod(\psi), \leq_\phi)$.
\label{proposition : AGM revision operator caracterization}
\end{theorem}

\begin{example}
Consider the propositional language defined from the set of atoms $\mathcal{A} = \{p, q\}$.
Let $\phi = p \Leftrightarrow \neg{q}$. Consider the total preorder $\leq_{\phi}$ defined as
$p \simeq_{\phi} q <_\phi pq <_\phi \emptyset$. It can be easily checked that the conditions of a faithful
assignment are satisfied by $\leq_\phi$. Then denote by $\circ$ the corresponding KM revision operator.
Now, let $\psi_1 = \neg{p} \wedge q$ and
$\psi_2 = p \Leftrightarrow q$.
Figure \ref{fig:preorderphi} illustrates the total preorder $\leq_{\phi}$ and graphically identifies the models of $\psi_1$ and $\psi_2$.
We get that:
\begin{itemize}
\item $mod(\phi \circ \psi_1) = \min(mod(\psi_1), \leq_\phi) = mod(\psi_1)$. Hence, $\phi \circ \psi_1 \equiv \psi_1$;
\item $mod(\phi \circ \psi_2) = \min(mod(\psi_2), \leq_\phi) = \{pq\}$. Hence, $\phi \circ \psi_2 \equiv p \wedge q$.
\end{itemize}
\begin{figure}[!htb]
	\centering
	\begin{tikzpicture}[yscale=.7]
		\tikzstyle{noeud}=[fill,circle,inner sep=1.5pt]
		\tikzstyle{niveau}=[-,line width=0.6pt]
		\tikzstyle{noeudlabel}=[node distance=0.3cm]
		\tikzstyle{flecheordre}=[->,dotted,line width=1pt,>=stealth]
		\tikzstyle{flechelabel}=[->,thin,>=stealth]
		\tikzstyle{textlabel}=[]
		\tikzstyle{cadrepsi}=[-, dashed, rounded corners=3pt]
		\tikzstyle{cadreresult}=[-, dashed, rounded corners=3pt, fill=gray!60, fill opacity=0.6]
		
		
		\fill[cadreresult] (3.7, 0.7) rectangle (4.3, 1.7);
		\draw[flechelabel] (5.5, 1.5) -- (4.32, 1.5);
		\draw[textlabel] (6.22, 1.52) node{$mod(\psi_1)$};
		
		\fill[cadreresult] (2.7, 1.7) rectangle (3.3, 3.7);
		\draw[flechelabel] (5.5, 3.5) -- (3.32, 3.5);
		\draw[textlabel] (6.25, 3.52) node{$mod(\psi_2)$};

		\draw[niveau] (1, 1) -- (5, 1);
		\draw[niveau] (1, 2) -- (5, 2);
		\draw[niveau] (1, 3) -- (5, 3);
		
		\draw[flecheordre] (.5, 1) -- (.5, 3);		
		
		\draw[textlabel] (-.02, 2) node{$\leq_\phi$};
		
		\node[noeud] (1) at (2, 1) {};
		\node[noeudlabel, above of=1] {$p$};
		\node[noeud] (2) at (4, 1) {};
		\node[noeudlabel, above of=2] {$q$};
		\node[noeud] (3) at (3, 2) {};
		\node[noeudlabel, above of=3] {$pq$};
		\node[noeud] (4) at (3, 3) {};
		\node[noeudlabel, above of=4] {$\emptyset$};
	\end{tikzpicture}
\caption{The total preorder $\leq_\phi$ over interpretations associated with some faithful assignment.}
\label{fig:preorderphi}
\end{figure}
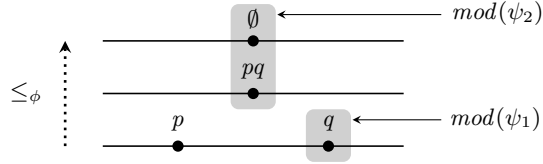
\label{ex:first one prop}
\end{example}

\nico{
In fact,
an implicit consequence of Theorem \ref{proposition : AGM revision operator caracterization} is that
every KM revision operator
is represented by a unique faithful assignment, and conversely, every faithful assignment represents a unique KM revision operator
(modulo equivalence):
\begin{proposition}
There is a one-to-one correspondence between the KM revision operators
and the set of all faithful assignments.
\label{rem:faithful}
\end{proposition}
}

KM revision operators include the class of distance-based revision operators (see, for instance, \cite{Dalal88}), i.e., those operators
characterized by a distance between interpretations:

\begin{definition}[Distance-based revision operators]
Let $d$ be a distance between interpretations\footnote{Actually, a pseudo-distance is enough, i.e.,
triangular inequality is not mandatory.}, extended to a distance between every interpretation $I$
and every formula $\phi$ by
$$d(I, \phi) =\left\lbrace\begin{array}{ll}
min\{d(I, J) \mid J \models \phi\} & \mbox{ if $\phi$ is consistent,}\\
0 \mbox{ otherwise.}
\end{array}\right.$$
The revision operator based on the distance $d$ is the operator $\circ^d$ satisfying for all formulae $\phi, \psi$,
$mod(\phi \circ^d \psi) = min(mod(\psi), \leq^d_\phi)$ where the preorder $\leq^d_\phi$
induced by $\phi$ is defined for all interpretations $I, J$ by $I \leq^d_\phi J$ if and only if $d(I, \phi) \leq d(J, \phi)$.
\label{def:distance-based revision operator}
\end{definition}
\nico{
The following result is a direct consequence of Theorem \ref{proposition : AGM revision operator caracterization}:
\begin{corollary}
Every distance-based revision operator is a KM revision operator, i.e.,
it satisfies the postulates (R1 - R6).
\end{corollary}
The result of revising old beliefs (a
propositional formula $\phi$) by new beliefs (a propositional formula $\psi$) is any propositional formula whose models are
models of $\psi$ having a distance to a model of $\phi$ which is minimal among all models of $\psi$.
}

\nico{
It is clear from Definition \ref{def:distance-based revision operator} that a distance fully characterizes the induced revision operator,
that is, different choices for the distance induce different revision operators.
}
Usual distances are $d_D$, the drastic distance ($d_D(I, J) = 1$ if and only if $I \neq J$), and $d_H$ the
Hamming distance ($d_H(I, J) = n$ if $I$ and $J$ differ on $n$ variables).
One can remark that when the drastic distance $d_D$ is used, the induced faithful assignment
associates with every formula $\phi$ a two-level preorder $\leq_\phi$;
\nico{
indeed, it can be easily verified that the revision operator
based on the drastic distance $d_D$ is equivalent to the so-called \textit{drastic revision operator}, which is defined
syntactically as follows:
\begin{definition}[Drastic revision operator]
The \emph{drastic revision operator}, denoted $\circ_D$, is the revision operator defined for all formulae
$\phi, \psi$ as
$$\phi \circ_D \psi =\left\lbrace\begin{array}{ll}
\phi \wedge \psi & \mbox{ if $\phi \wedge \psi$ is consistent,}\\
\psi & \mbox{ otherwise.}
\end{array}\right.$$
\label{def:drasticr}
\end{definition}
This operator was first introduced in \cite{AGM85} under the name of \textit{full meet revision function}.
Though ``fully rational'' in the sense that it satisfies all the KM rationality postulates (i.e., all AGM postulates in \cite{AGM85}),
it is often considered as unreasonable because it throws away all the old beliefs if the new formula is inconsistent with them.
}



\nico{
Likewise, the revision operator based on Hamming distance $d_H$ is equivalent to the well-known
Dalal revision operator \cite{Dalal88}. In fact, in \cite{Dalal88} the Dalal revision
is also defined in a modelwise fashion, i.e., there is no syntactic definition of it (as opposite
to the drastic revision operator, cf. Definition \ref{def:drasticr}):
\begin{definition}[Dalal revision operator]
A \emph{Dalal revision operator}, denoted $\circ_{Dal}$, is any revision operator
based on the Hamming distance.
\label{def:dalal}
\end{definition}
From now on, the revision operator
based on the Hamming distance (i.e., the revision operator $\circ^{d_H}$) will simply be referred as the Dalal revision operator,
and thus will be denoted $\circ_{Dal}$.
}

\nico{
\begin{example}
Let
$\mathcal{A} = \{p, q, r\}$,
$\phi = p \wedge q \wedge \neg{r}$, and $\psi = r$.
We have
\begin{itemize}
\item $\phi \circ_D \psi = r$.
\item $\phi \circ_{Dal} \psi \equiv p \wedge q \wedge r$.
\end{itemize}
\label{ex:drast-dalal}
\end{example}
}

\nico{
It is clear from Example \ref{ex:drast-dalal} that the Dalal revision operator has a more parsimonious behavior than the drastic
revision operator, because it integrates the new information while keeping as much previous beliefs as possible.
}

\nico{
Before concluding this section, let us remark that distance-based revision operators as defined above do not
fully characterize KM revision operators: this comes from the fact that given two formulae $\phi, \phi'$
such that $\phi \not\equiv \phi'$, one can associate within the same faithful assignment two preorders
$\leq_\phi, \leq_{\phi'}$ in an independent way; given that observation, one can easily build $\leq_\phi, \leq_{\phi'}$
using two different distances, whereas Definition \ref{def:distance-based revision operator} requires that the \emph{same} distance
is used to define the total preorder $\leq_\phi$ associated with \emph{any} formula. However, as far as we know there does not exist
in the literature any ``fully rational'' (with respect to postulates (R1 - R6)) revision operator of interest that is not distance-based.
}

\section{Belief revision in Logic Programming}
\label{sec:belief revision logic programming}
\subsection{Preliminaries on Logic Programming}

We define the syntax and semantics of generalized
logic programs.
We use the same notations as in \cite{DBLP:conf/kr/DelgrandeSTW08}.
A \textit{generalized logic program} (GLP) is a finite set of rules of the form

$$\begin{array}{l}
a_1; \dots; a_k; \sim b_1; \dots; \sim b_l \leftarrow c_1, \dots, c_m, \sim d_1, \dots, \sim d_n,
\end{array}\trichea$$
where $k, l, m, n \geq 0$.

Each $a_i, b_i, c_i, d_i$ is either one of the constant symbols $\bot$, $\top$, or an atom from $\mathcal{A}$;
$\sim$ is the negation by failure; ``$;$'' is the disjunctive connective, ``$,$'' is the conjunctive connective of atoms.
The right-hand and left-hand
sides of $r$ are respectively called the head and body of $r$.
For each rule $r$, we define $H(r)^{+} = \{a_1, \dots, a_k\}$, $H(r)^{-} = \{b_1, \dots, b_l\}$, $B(r)^{+} = \{c_1, \dots, c_m\}$, and $B(r)^{-} = \{d_1, \dots,$ $d_n\}$.
For the sake of simplicity, a rule $r$ is also expressed as follows:

$$H(r)^{+} ; \sim H(r)^{-} \leftarrow B(r)^{+} , \sim B(r)^{-}.$$

A logic program is interpreted through its preferred models based on the answer set semantics.
A \textit{(classical) model} $X$ of a GLP $\P$ (written $X \models \P$) is an interpretation from $\allinterpretations$ that satisfies all rules from $\P$
according to the classical definition of truth in propositional logic. $mod(\P)$ will denote the set of all models of a GLP $\P$.
An \emph{answer set} $X$ of a GLP $\P$ is a minimal (w.r.t. set inclusion) set of atoms from $\mathcal{A}$ that is a model of the program $\P^X$, where
$\P^X$ is called the \emph{reduct} of $\P$ relative to $X$ and is defined as
$\P^X = \{H(r)^{+} \leftarrow B(r)^{+} \mid r \in \P, H(r)^{-} \subseteq X, B(r)^{-} \cap X = \emptyset\}$.
The classical notion of equivalence between programs corresponds to the correspondence of their answer sets.
Recall that we denote an interpretation by dropping set-notations except for the case of the interpretation corresponding
to the empty set; for instance, the set of interpretations
$\{\emptyset, \{p\}, \{pq\}\}$ will be simply denoted $\{\emptyset, p, pq\}$.
\newpage

\begin{example}
Consider the logic program $\P =
\left\lbrace\begin{array}{l}
p \leftarrow \sim q,\\
\bot \leftarrow p, q
\end{array}\right\rbrace$.
To determine $AS(\P)$, the set of answer sets of $\P$, we need to check for each interpretation $X$ whether $X$ is
a minimal (w.r.t. set inclusion) model of $\P^X$, the reduct of $\P$ relative to $X$:
\begin{itemize}
\item $\P^\emptyset =
\left\lbrace\begin{array}{l}
p \leftarrow \top,\\
\bot \leftarrow p, q
\end{array}\right\rbrace$,
and $mod(\P^\emptyset) = \{p\}$. Since $\emptyset$ is not
a model of $\P^\emptyset$, we get that $\emptyset \notin AS(\P)$;
\item $\P^{p} = \P^\emptyset$, so $mod(\P^{p}) = \{p\}$. Since $p$ is a minimal
(w.r.t. set inclusion) model of $\P^{p}$, we get that $p \in AS(\P)$;
\item $\P^{q} = \{\bot \leftarrow p, q\}$, so $mod(\P^{q}) = \{\emptyset, p, q\}$. Hence, $q$ is a
model of $\P^{q}$ but is not minimal w.r.t. set inclusion, since $\emptyset \in mod(\P^{q})$. Thus $q \notin AS(\P)$;
\item lastly, $\P^{pq} = \P^{q}$, so $mod(\P^{pq}) = \{\emptyset, p, q\}$.
Hence, $pq$ is a
not a model of $\P^{pq}$, so we get that $pq \notin AS(\P)$;
\end{itemize}
Therefore, $AS(\P) = \{p\}$.
\label{ex:detailed GLP P}
\end{example}



\emph{SE interpretations} 
are semantic structures characterizing \emph{strong equivalence} between logic programs \cite{DBLP:journals/tplp/Turner03},
they provide a monotonic semantic foundation of logic programs under answer set semantics.
An SE interpretation over $\mathcal{A}$ is a pair $(X, Y)$ of interpretations over $\mathcal{A}$ such that $X \subseteq Y$.
An \emph{SE model} $(X, Y)$ of a logic program $\P$ is an SE interpretation over $\mathcal{A}$ that satisfies $Y \models \P$ and $X \models \P^Y$, where $\P^Y$ is
the reduct of $\P$ relative to $Y$. The set $\SE$ denotes the set of all SE interpretations over $\mathcal{A}$; given a logic program $\P$, the set $\SE(\P)$ denotes the set of
SE models of $\P$.

\begin{example}
Consider again the logic program $\P$ defined in Example \ref{ex:detailed GLP P}. We have $mod(\P) = \{p, q\}$. Hence,
$$\begin{array}{ll}
\SE(\P) & = \{(X, p) \in \SE \mid X \in mod(\P^{p})\} \cup \{(X, q) \in \SE \mid X \in mod(\P^{q})\}\\
& = \{(X, p) \in \SE \mid X \in \{p\}\} \cup \{(X, q) \in \SE \mid X \in \{\emptyset, p, q\}\}\\
& = \{(p, p), (\emptyset, q), (q, q)\}.
\end{array}$$
\end{example}

Through their SE models, logic programs are semantically described in
a stronger way than through their answer sets, as shown in the following example.

\begin{example}
Let $\P_1 = \{p \leftarrow \sim q\}$ and $\P_2 =
\left\lbrace\begin{array}{l}
p \leftarrow \sim{q},\\
p; q \leftarrow \top
\end{array}\right\rbrace$, and consider again the logic program $\P$ defined in Example \ref{ex:detailed GLP P}.
Then we get that
$$AS(\P) = AS(\P_1) = AS(\P_2) = \{p\},$$ that is, $\P$, $\P_1$ and $\P_2$ admit the same answer sets. However their SE models differ:
$$\begin{array}{ll}
\SE(\P) = \{(p, p), (\emptyset, q), (q, q)\} & \mbox{ (cf. Example \ref{ex:detailed GLP P})},\\
\SE(\P_1) = \{(p, p), (\emptyset, q), (q, q), (\emptyset, pq), (p, pq), (q, pq), (pq, pq)\}, & \\
\SE(\P_2) = \{(p, p), (p, pq), (q, pq), (pq, pq)\}, &
\end{array}$$
\label{ex:difference-asp-se}
\end{example}

A program $\P$ is \emph{consistent} if $\SE(P) \neq \emptyset$.
Two programs $\P$ and $\Q$ are said to be \emph{strongly equivalent}, denoted $\P \equiv_s \Q$, whenever $\SE(\P) = \SE(\Q)$.
We also write $\P \subseteq_s \Q$ if $\SE(\P) \subseteq \SE(\Q)$. Two programs are equivalent if they are strongly equivalent,
but the other direction does not hold in general (cf. Example \ref{ex:difference-asp-se}).
Note that $Y$ is an answer set of $\P$ if and only if $(Y, Y) \in \SE(\P)$ and no $(X, Y) \in \SE(\P)$
with $X \subsetneq Y$ exists. We also have $(Y, Y) \in \SE(\P)$ if and only if $Y \in mod(\P)$.
A set of SE interpretations $S$ is \emph{well-defined} if for every interpretation $X, Y$ with $X \subseteq Y$, if $(X, Y) \in S$ then $(Y, Y) \in S$.
Every GLP
has a well-defined
set of SE models.
Moreover, from every well-defined
set $S$ of SE models, one can build a GLP
$P$ such that $\SE(P) = S$ \cite{Eiter05onsolution,Cabalar}.

We close this section by introducing two further notations.
For every GLP $\P$, $\formGLPb{\P}$ is any propositional formula satisfying
$\modeles{\formGLPb{\P}} = \modeles{\P}$, and $\formGLPa{\P}$ is any propositional
formula satisfying $\modeles{\formGLPa{\P}} = \{X \in \allinterpretations \mid (X, Y) \in \SE(\P)\}$.

\subsection{Logic program revision operators}
\label{sec:LP revision operators}
We now consider belief revision in the context of logic programs. Given two programs $\P, \Q$
the goal is to define a program $\P \star \Q$ which is the revision of $\P$ by $\Q$.
Delgrande \textit{et al.\/} \citeyear{DBLP:conf/kr/DelgrandeSTW08,Delgrande/ACM}
proposed an adaptation of the KM postulates (cf. Definition \ref{def:KM revision operator})
in the context of logic programming; this can be done using the
monotonic characterization of logic programs through their SE models.
First, they considered the operation of \textit{expansion} of two logic programs:

\begin{definition}[Expansion operator \cite{DBLP:conf/kr/DelgrandeSTW08}]
Given two programs $\P, \Q$, the \emph{expansion} of $\P$ by $\Q$, denoted $\P + \Q$ is any program $\R$ such that
$\SE(\R) = \SE(\P) \cap \SE(\Q)$.
\label{def:expansion}
\end{definition}

Though the expansion of logic programs
trivializes the result whenever the two input logic programs admit no common SE models,
this operation is of interest in its own right.
For instance, it can be observed that the intersection of two well-defined sets of SE interpretations
leads to a well-defined set of SE interpretations, and thus the expansion of two GLPs is always defined as a GLP.

\begin{example}
Consider again the program $\P$ from Example \ref{ex:detailed GLP P}, and recall that $\SE(\P) = \{(p, p),$ \linebreak
$(\emptyset, q), (q, q)\}$.
Let $\Q$ be the GLP $\Q = \{q \leftarrow \top\}$, we have $\SE(\Q) = \{(q, q), (q, pq),$ \linebreak $(pq, pq)\}$.
Furthermore, the GLP $\R =
\left\lbrace\begin{array}{l}
q \leftarrow \top,\\
\bot \leftarrow p
\end{array}\right\rbrace$
is such that $\SE(\R) = \{(q, q)\} = \SE(\P) \cap \SE(\Q)$.
Therefore,
$$\P + \Q = \left\lbrace\begin{array}{l}
p \leftarrow \sim q,\\
\bot \leftarrow p, q
\end{array}\right\rbrace + \{q \leftarrow \top\} \equiv_s \left\lbrace\begin{array}{l}
q \leftarrow \top,\\
\bot \leftarrow p
\end{array}\right\rbrace.$$
\label{ex:expansion}
\end{example}

We refer the reader to \cite{Delgrande/ACM}, Section 3.1 for further examples of the use of the expansion operator.


%

Expansion of programs corresponds to the model-theoretical definition of expansion expressed through the KM postulates
R2, R5 and R6.
Delgrande \textit{et al.\/} rephrased the full set of KM postulates (R1 - R6) in the context of
GLPs.
Beforehand, we define a logic program revision operator as a simple function, that considers
two GLPs (the original one and the new one) and returns a revised GLP:

\nico{
\begin{definition}[LP revision operator, equivalence between LP revision operators]
A \emph{LP revision operator} $\star$ is a mapping associating two GLPs $\P, \Q$ with
a new GLP, denoted $\P \star \Q$.
Two LP revision operators $\star, \star'$ are said to be \emph{equivalent} (denoted $\star \equiv \star'$) when for all GLPs $\P, \Q$,
$\P \star \Q \equiv_s \P \star' \Q$.
\label{def:LP rev op}
\end{definition}
}

\begin{definition}[GLP revision operator \cite{DBLP:conf/kr/DelgrandeSTW08}]
A \emph{GLP revision operator} $\revision$ is an LP revision operator that satisfies the following postulates,
for all GLPs $\P, \P_1, \P_2, \Q, \Q_1, \Q_2, \R$:
\begin{description}
	\item[\textbf{(RA1)}] $\P \revision \Q \subseteq_s \Q$;
	\item[\textbf{(RA2)}] If $\P + \Q$ is consistent, then $\P \revision \Q \equiv_s \P + \Q$;
	\item[\textbf{(RA3)}] If $\Q$ is consistent, then $\P \revision \Q$ is consistent;
	\item[\textbf{(RA4)}] If $\P_1 \equiv_s \P_2$ and $\Q_1 \equiv_s \Q_2$, then $\P_1 \revision \Q_1 \equiv_s \P_2 \revision \Q_2$;
	\item[\textbf{(RA5)}] $(\P \revision \Q) + \R \subseteq_s \P \revision (\Q + \R)$;
	\item[\textbf{(RA6)}] If $(\P \revision \Q) + \R$ is consistent, then $\P \revision (\Q + \R) \subseteq_s (\P \revision \Q) + \R$.
\end{description}
\label{def:GLP-revision-operator}
\end{definition}




%
%

\nico{
As to the case of (propositional) KM revision operators, 
an LP revision operator $\star$ is a GLP revision operator if and only if
any LP revision operator equivalent to $\star$ is also a GLP revision operator.
This is why
in the rest of the paper,
as we identify a propositional revision operator modulo equivalence,
we also identify an LP revision operator modulo equivalence. This allows us to
define a revised program in a modelwise fashion, i.e., as its set of SE models,
and given two LP revision operators $\star, \star'$, the notations $\star \equiv \star'$ and
$\star = \star'$ are confunded with no harm.
}


Delgrande \textit{et al.\/} \citeyear{DBLP:conf/kr/DelgrandeSTW08} proposed
a revision operator
inspired from
Satoh's propositional revision operator \cite{DBLP:conf/fgcs/Satoh88}. This operator, based on the set containment of SE interpretations,
satisfies postulates (RA1 - RA5). Though it seems to have a good
behavior on some instances, this operator does not satisfy (RA6), so that it does
not fully respect the principle of minimality of change (see \cite{DBLP:conf/ijcai/KatsunoM89}, Section 3.1
for details on this postulate). However, the whole set of postulates is
consistent, as they later introduce the so-called
\emph{cardinality-based revision operator} \cite{Delgrande/ACM} that reduces
to the Dalal revision scheme over propositional models and that satisfies all the postulates (RA1 - RA6).
The following definition is a concise, equivalent reformulation of the original one introduced in \cite{Delgrande/ACM},
Definition 3.10:

\begin{definition}[Cardinality-based revision operator]
Given a GLP $\P$ and an interpretation $Y$, let
$form_Y$ be any propositional formula satisfying $mod(form_Y) = \{Y\}$,
let $\alpha_{(\P, Y)}$ be any propositional formula satisfying
$mod(\alpha_{(\P, Y)}) = \{X \in \Omega \mid (X, Y') \in \SE(\P), Y' \models form_Y \circ_{Dal} \alpha^2_\P \}$,
and let $\alpha_{Y}$ be any propositional formula satisfying
$mod(\alpha_{Y}) = \{X \in \Omega \mid X \subseteq Y\}$.
The cardinality-based revision operator, denoted $\star_c$, is defined for all
GLPs $\P, \Q$ by any program $\P \star_c \Q$ satisfying
$$\begin{array}{l}
\SE(\P \star_c \Q) = \{(X, Y) \in \SE(\Q) \mid Y \models \formGLPb{\P} \circ_{Dal} \formGLPb{\Q}\\
\ \ \ \ \ \ \ \ \ \ \ \ \ \ \ \ \ \ \ \ \ \ \ \ \mbox{ and if } X \subsetneq Y \mbox{ then } X \models \alpha_{(\P, Y)} \circ_{Dal} \alpha_Y\}\}.
\end{array}$$
\label{def:cardinal-based}
\end{definition}

\begin{theorem}[Delgrande et al. 2013b]
$\star_c$ is a GLP revision operator.
\label{th:cardinal-based revision operator}
\end{theorem}

In addition, we introduce below a simple, syntactically defined LP revision operator which also satisfies the whole set of postulates
(RA1 - RA6):



\begin{definition}[Drastic LP revision operator]
The \emph{drastic LP revision operator} $\revision_D$ is defined for all GLPs $\P, \Q$ as
$$\P \revision_D \Q =\left\lbrace\begin{array}{ll}
\P + \Q & \mbox{ if $\P + \Q$ is consistent,}\\
\Q & \mbox{ otherwise.}
\end{array}\right.$$
\label{def:drastic}
\end{definition}

\begin{proposition}
$\revision_D$ is a GLP revision operator.
\label{prop:drastic revision operator}
\end{proposition}

Note that the drastic LP revision operator is the counterpart of the propositional drastic revision operator
(cf. Definition \ref{def:drasticr}) for logic programs: the old program is thrown away if the new program is inconsistent
with it. The cardinality-based revision operator has a more parsimonious behavior.
However, Theorem \ref{th:cardinal-based revision operator} and Proposition \ref{prop:drastic revision operator}
show that these operators
are both fully satisfactory in terms of revision principles; this raises the problem
on how to discard some rational operators from others.
Moreover, it is not clear whether there even exist other GLP revision operators than the cardinality-based
and the drastic LP revision operators.
In the next section, we fill the gap and we give a constructive, full characterization of
the class of GLP revision operators, that provides us a clear and complete picture of it.

\section{Characterization of GLP revision operators}
\label{sec:representation}
\subsection{Characterization result}
We now provide the main result of our paper, i.e., a characterization theorem for GLP revision operators.
That is,
we show that each GLP revision operator (i.e., each LP revision operator satisfying the postulates
(RA1 - RA6)) can be characterized in terms of preorders
over the set of all classical interpretations, with some further conditions specific to SE interpretations.


\begin{definition}[LP faithful assignment]
An \emph{LP faithful assignment} is a mapping which associates with every GLP $\P$
a total preorder $\leq_\P$ over interpretations such that for all GLPs $\P, \Q$ and
all interpretations $Y, Y'$, the following conditions hold:
\begin{description}
\item[(1)] If $Y \models \P$ and $Y' \models \P$, then $Y \simeq_\P Y'$;
\item[(2)] If $Y \models \P$ and $Y' \not\models \P$, then $Y <_\P Y'$;
\item[(3)] If $\modeles{\P} = \modeles{\Q}$, then $\leq_\P = \leq_\Q$.
\end{description}
\label{def:LPfaithful}
\end{definition}

Please note the similarities between an LP faithful assignment and a faithful assignment
(cf. Definition \ref{def:faithful}). That is:

\begin{remark}
Let $\Phi_1$ be an assignment that associates with
every GLP $\P$ a total preorder $\leq_\P$ over interpretations, and $\Phi_2$ be
and assignment that
associates with every formula $\phi$ a total preorder $\leq_\phi$ over interpretations. If
for every GLP $\P$, we have $\Phi_1(\P) = \Phi_2(\formGLPb{\P})$, then
$\Phi_1$ is an LP faithful assignment if and only if $\Phi_2$ is a faithful assignment.
\label{remark}
\end{remark}

\begin{definition}[Well-defined assignment]
A \emph{well-defined assignment} is a
mapping which associates with every GLP $\P$
and every interpretation $Y$ a set of interpretations, denoted by $\P(Y)$,
such that for all GLPs $\P, \Q$ and all interpretations
$X, Y$, the following conditions hold:
\begin{description}
\item[(a)] $Y \in \P(Y)$;
\item[(b)] If $X \in \P(Y)$, then $X \subseteq Y$;
\item[(c)] If $(X, Y) \in \SE(\P)$, then $X \in \P(Y)$;
\item[(d)] If $(X, Y) \notin \SE(\P)$ and $Y \models \P$, then $X \notin \P(Y)$;
\item[(e)] If $\P \equiv_s \Q$, then $\P(Y) = \Q(Y)$.
\end{description}
\label{def:well-defined-assignment}
\end{definition}



\begin{definition}[GLP parted assignment]
A \emph{GLP parted assignment} is a pair $(\Phi, \Psi)$, where
$\Phi$ is an LP faithful assignment and $\Psi$ is a well-defined assignment.
\label{def:parted-assignment}
\end{definition}

We are ready to bring to light our main result:


\begin{proposition}
An LP operator $\star$ is a GLP revision operator if and only if there exists a
GLP parted assignment $(\Phi, \Psi)$, where
$\Phi$ associates with every GLP $\P$ a total preorder $\leq_\P$,
$\Psi$ associates with every GLP $\P$ and every interpretation $Y$ a set of interpretations $\P(Y)$, and
such that for all GLPs $\P, \Q$,
$$\SE(\P \star \Q) = \{(X, Y) \mid (X, Y) \in \SE(\Q),
Y \in \min(\modeles{\Q}, \leq_\P),
X \in \P(Y)\}.$$
\label{prop:characterization-revision-GLP}
\end{proposition}

%


Note that there is no relationship between the LP faithful assignment $\Phi$ and the well-defined assignment $\Psi$
forming a GLP parted assignment,
that is, each one of these two mappings can be defined in a completely independent way.

\begin{example}
Let us consider again the GLP $\P =
\left\lbrace\begin{array}{l}
p \leftarrow \sim q,\\
\bot \leftarrow p, q
\end{array}\right\rbrace$ from Example \ref{ex:detailed GLP P}, and recall that
$\SE(\P) = \{(p, p), (\emptyset, q), (q, q)\}$.
Note that the (classical) models of $\P$ (i.e., $mod(\P) = \{p, q\}$) correspond to the models of the propositional formula $\phi$ given in
Example \nolinebreak\ref{ex:first one prop} (i.e., $mod(\phi) = \{p, q\}$). Hence, due to Remark \ref{remark} the total preorder $\leq_\P = \leq_\phi$, i.e., defined as
$p \simeq_\P q <_\P pq <_\P \emptyset$ satisfies the conditions of an LP faithful
assignment (denoted $\Phi$). Furthermore, let us consider the mapping $\Psi$ associating with $\P$ and every
interpretation $Y$ the following sets of interpretations:
$\P(\emptyset) = \{\emptyset\}$, $\P(p) = \{p\}$, $\P(q) = \{\emptyset, q\}$ and $\P(pq) = \{p, pq\}$.
One can also check that $\Psi$ satisfies the conditions (a - e) from Definition \ref{def:well-defined-assignment}, so $\Psi$
is a well-defined assignment. Hence, $(\Phi, \Psi)$ is a GLP parted assignment. Figure \ref{fig:GLP-parted-assignment1}
gives a graphical representation of the total preorder $\leq_\P$ and the sets $\P(Y)$ for each $Y \in \Omega$.
In the figure, all interpretations are ordered w.r.t. $\leq_\P$ (similarly to Figure \ref{fig:preorderphi}), and for each
such interpretation $Y$, the set of circle interpretations next to $Y$ corresponds to the set $\P(Y)$.
\begin{figure}[!htb]
	\centering
		\begin{tikzpicture}[yscale=1.2,xscale=1.2]
		\tikzstyle{noeud}=[fill,circle,inner sep=1.5pt]
		\tikzstyle{niveau}=[-,line width=0.6pt]
		\tikzstyle{noeudlabel}=[node distance=0.3cm]
		\tikzstyle{flecheordre}=[->,dotted,line width=1pt,>=stealth]
		\tikzstyle{flechelabel}=[->,thin,>=stealth]
		\tikzstyle{textlabel}=[]
		\tikzstyle{cadrepsi}=[-, dashed, rounded corners=3pt]
		\tikzstyle{cadreresult}=[-, dashed, rounded corners=2pt, fill=gray!60, fill opacity=0.6]
		\tikzstyle{cadrewelldefined}=[-, very thin, rounded corners=3pt, dotted, fill=gray!60, fill opacity=0.15]
		\tikzstyle{linewelldefined}=[-, very thin]
		\tikzstyle{flechewelldefined}=[->,thin,>=stealth]
		\tikzstyle{cadrePof}=[-, rounded corners=2pt]
		
		
		
%
		
		\fill[cadrewelldefined] (2.7, 1.1) rectangle (3.8, 1.8);
		\draw[textlabel] (2.95, 1.63) node{$\emptyset$};
		\draw[textlabel] (3.45, 1.61) node{$p$};
		\draw[textlabel] (2.95, 1.26) node{$q$};
		\draw[textlabel] (3.45, 1.26) node{$pq$};
		\path[flechewelldefined] (2.1, 1.45) edge[out=50,in=190] (2.68, 1.7);	
		\draw[textlabel] (2.2, 1.76) node{\tiny{$\P(p)$}};
		
		\draw[cadrePof] (3.25, 1.48) rectangle (3.65, 1.75);
		
		\fill[cadrewelldefined] (4.9, 1.1) rectangle (6, 1.8);
		\draw[textlabel] (5.15, 1.63) node{$\emptyset$};
		\draw[textlabel] (5.65, 1.61) node{$p$};
		\draw[textlabel] (5.15, 1.26) node{$q$};
		\draw[textlabel] (5.65, 1.26) node{$pq$};
		\path[flechewelldefined] (4.3, 1.45) edge[out=50,in=190] (4.88, 1.7);	
		\draw[textlabel] (4.4, 1.76) node{\tiny{$\P(q)$}};
		
		\draw[cadrePof] (4.95, 1.12) rectangle (5.35, 1.765);
		
		\fill[cadrewelldefined] (3.8, 2.1) rectangle (4.9, 2.8);
		\draw[textlabel] (4.05, 2.63) node{$\emptyset$};
		\draw[textlabel] (4.55, 2.61) node{$p$};
		\draw[textlabel] (4.05, 2.26) node{$q$};
		\draw[textlabel] (4.55, 2.26) node{$pq$};
		\path[flechewelldefined] (3.2, 2.45) edge[out=50,in=190] (3.78, 2.7);	
		\draw[textlabel] (3.3, 2.76) node{\tiny{$\P(pq)$}};
		
		\draw[cadrePof] (4.35, 2.12) rectangle (4.75, 2.75);
		
		\fill[cadrewelldefined] (3.8, 3.1) rectangle (4.9, 3.8);
		\draw[textlabel] (4.05, 3.63) node{$\emptyset$};
		\draw[textlabel] (4.55, 3.61) node{$p$};
		\draw[textlabel] (4.05, 3.26) node{$q$};
		\draw[textlabel] (4.55, 3.26) node{$pq$};
		\path[flechewelldefined] (3.2, 3.45) edge[out=50,in=190] (3.78, 3.7);	
		\draw[textlabel] (3.3, 3.76) node{\tiny{$\P(\emptyset)$}};
		
		\draw[cadrePof] (3.85, 3.48) rectangle (4.25, 3.765);		
		
		\draw[niveau] (1, 1) -- (5, 1);
		\draw[niveau] (1, 2) -- (5, 2);
		\draw[niveau] (1, 3) -- (5, 3);
		
		\draw[flecheordre] (0.5, 1) -- (0.5, 3);		
		
		\draw[textlabel] (0.2, 2) node{$\leq_\P$};
		
		\node[noeud] (1) at (2, 1) {};
		\node[noeudlabel, above of=1] {$p$};
		\node[noeud] (2) at (4.2, 1) {};
		\node[noeudlabel, above of=2] {$q$};
		\node[noeud] (3) at (3.1, 2) {};
		\node[noeudlabel, above of=3] {$pq$};
		\node[noeud] (4) at (3.1, 3) {};
		\node[noeudlabel, above of=4] {$\emptyset$};
	\end{tikzpicture}
\caption{The total preorder $\leq_\P$ over SE interpretations, and the sets $\P(Y)$ enclosed in boxes for all $Y \in \Omega$, associated with some GLP parted assignment.}
\label{fig:GLP-parted-assignment1}
\end{figure}
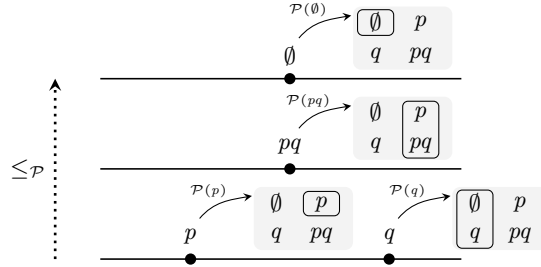

Now, let us denote $\star$ the GLP revision operator corresponding to this GLP parted assignment, and let $\Q_1$ and $\Q_2$ be two GLPs
defined as $\Q_1 = \{q \leftarrow \sim{p}\}$ and \linebreak $\Q_2 =
\left\lbrace\begin{array}{ll}
\bot \leftarrow p, \sim q, & \bot \leftarrow q, \sim p,\\
p; \sim{p} \leftarrow \top, & q; \sim{q} \leftarrow \top.
\end{array}\right\rbrace.$
We get that:
\begin{itemize}
\item $\SE(\Q_1) = \{(\emptyset, p), (p, p), (q, q), (\emptyset, pq), (p, pq), (q, pq), (pq, pq)\}$; then according to
Proposition \ref{prop:characterization-revision-GLP}, we get that
$\SE(\P \star \Q_1) = \{(p, p), (q, q)\}$. Furthermore, the GLP
$\R_1 =
\left\lbrace\begin{array}{l}
p \leftarrow \sim q,\\
q \leftarrow \sim p,\\
\bot \leftarrow p, q
\end{array}\right\rbrace$
is such that
$\SE(\R_1) = \{(p, p), (q, q)\} = \SE(\P \star \Q_1)$.
Therefore,
$$\P \star \Q_1 = \left\lbrace\begin{array}{l}
p \leftarrow \sim q,\\
\bot \leftarrow p, q
\end{array}\right\rbrace \star
\{q \leftarrow \sim{p}\}
\equiv_s
\left\lbrace\begin{array}{l}
p \leftarrow \sim q,\\
q \leftarrow \sim p,\\
\bot \leftarrow p, q
\end{array}\right\rbrace.$$
\item $\SE(\Q_2) = \{(\emptyset, \emptyset), (pq, pq)\}$; then according to
Proposition \ref{prop:characterization-revision-GLP}, we get that
$\SE(\P \star \Q_2) = \{(pq, pq)\}$. Furthermore, the GLP $\R_2 =
\left\lbrace\begin{array}{l}
p \leftarrow \top,\\
q \leftarrow \top
\end{array}\right\rbrace$ is such that
$\SE(\R_2) = \{(pq, pq)\} = \SE(\P \star \Q_2)$.
Therefore,
$$\P \star \Q_2 = \left\lbrace\begin{array}{l}
p \leftarrow \sim q,\\
\bot \leftarrow p, q
\end{array}\right\rbrace \star
\left\lbrace\begin{array}{ll}
\bot \leftarrow p, \sim q, & \bot \leftarrow q, \sim p,\\
p; \sim{p} \leftarrow \top, & q; \sim{q} \leftarrow \top.
\end{array}\right\rbrace
\equiv_s
\left\lbrace\begin{array}{l}
p \leftarrow \top,\\
q \leftarrow \top
\end{array}\right\rbrace.$$
\end{itemize}
The SE models of $\Q_1$ and $\Q_2$ are respectively illustrated in Figures \ref{subfig-GLPa} and \ref{subfig-GLPb}.
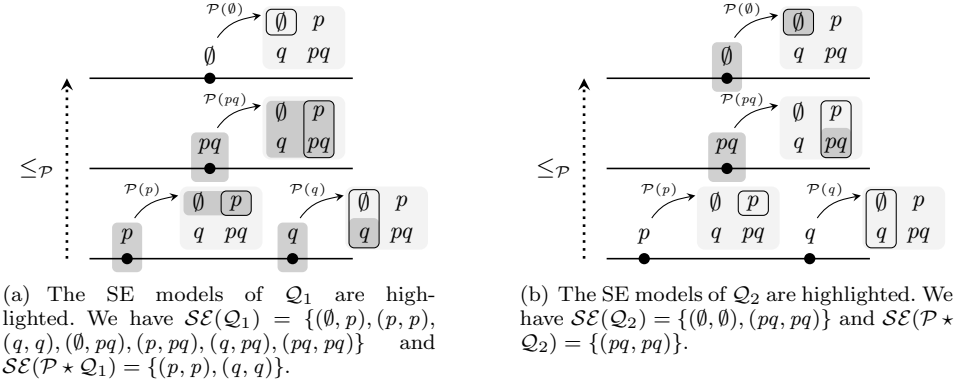
\begin{figure}[!htb]
	\centering
	\subfigure[The SE models of $\Q_1$ are highlighted. We have $\SE(\Q_1) = \{(\emptyset, p), (p, p),$ $(q, q), (\emptyset, pq), (p, pq), (q, pq), (pq, pq)\}$
	and $\SE(\P \star \Q_1) = \{(p, p), (q, q)\}$.]{\label{subfig-GLPa}
		\begin{tikzpicture}[yscale=1.2,xscale=1]
		\tikzstyle{noeud}=[fill,circle,inner sep=1.5pt]
		\tikzstyle{niveau}=[-,line width=0.6pt]
		\tikzstyle{noeudlabel}=[node distance=0.3cm]
		\tikzstyle{flecheordre}=[->,dotted,line width=1pt,>=stealth]
		\tikzstyle{flechelabel}=[->,thin,>=stealth]
		\tikzstyle{textlabel}=[]
		\tikzstyle{cadrepsi}=[-, dashed, rounded corners=3pt]
		\tikzstyle{cadreresult}=[-, dashed, rounded corners=2pt, fill=gray!60, fill opacity=0.6]
		\tikzstyle{cadrewelldefined}=[-, very thin, rounded corners=3pt, dotted, fill=gray!60, fill opacity=0.15]
		\tikzstyle{linewelldefined}=[-, very thin]
		\tikzstyle{flechewelldefined}=[->,thin,>=stealth]
		\tikzstyle{cadrePof}=[-, rounded corners=2pt]
		
		
		\fill[cadreresult] (1.8, 0.85) rectangle (2.2, 1.4);
		\fill[cadreresult] (2.75, 1.48) rectangle (3.65, 1.75);
		\fill[cadreresult] (4, 0.85) rectangle (4.4, 1.4);
		\fill[cadreresult] (4.95, 1.12) rectangle (5.35, 1.45);
		\fill[cadreresult] (2.85, 1.85) rectangle (3.35, 2.4);
		\fill[cadreresult] (3.85, 2.12) rectangle (4.75, 2.75);
		
%
		
		\fill[cadrewelldefined] (2.7, 1.1) rectangle (3.8, 1.8);
		\draw[textlabel] (2.95, 1.63) node{$\emptyset$};
		\draw[textlabel] (3.45, 1.61) node{$p$};
		\draw[textlabel] (2.95, 1.26) node{$q$};
		\draw[textlabel] (3.45, 1.26) node{$pq$};
		\path[flechewelldefined] (2.1, 1.45) edge[out=50,in=190] (2.68, 1.7);
		\draw[textlabel] (2.2, 1.76) node{\tiny{$\P(p)$}};
		
		\draw[cadrePof] (3.25, 1.48) rectangle (3.65, 1.75);
		
		\fill[cadrewelldefined] (4.9, 1.1) rectangle (6, 1.8);
		\draw[textlabel] (5.15, 1.63) node{$\emptyset$};
		\draw[textlabel] (5.65, 1.61) node{$p$};
		\draw[textlabel] (5.15, 1.26) node{$q$};
		\draw[textlabel] (5.65, 1.26) node{$pq$};
		\path[flechewelldefined] (4.3, 1.45) edge[out=50,in=190] (4.88, 1.7);	
		\draw[textlabel] (4.4, 1.76) node{\tiny{$\P(q)$}};
		
		\draw[cadrePof] (4.95, 1.12) rectangle (5.35, 1.765);
		
		\fill[cadrewelldefined] (3.8, 2.1) rectangle (4.9, 2.8);
		\draw[textlabel] (4.05, 2.63) node{$\emptyset$};
		\draw[textlabel] (4.55, 2.61) node{$p$};
		\draw[textlabel] (4.05, 2.26) node{$q$};
		\draw[textlabel] (4.55, 2.26) node{$pq$};
		\path[flechewelldefined] (3.2, 2.45) edge[out=50,in=190] (3.78, 2.7);	
		\draw[textlabel] (3.3, 2.76) node{\tiny{$\P(pq)$}};
		
		\draw[cadrePof] (4.35, 2.12) rectangle (4.75, 2.75);
		
		\fill[cadrewelldefined] (3.8, 3.1) rectangle (4.9, 3.8);
		\draw[textlabel] (4.05, 3.63) node{$\emptyset$};
		\draw[textlabel] (4.55, 3.61) node{$p$};
		\draw[textlabel] (4.05, 3.26) node{$q$};
		\draw[textlabel] (4.55, 3.26) node{$pq$};
		\path[flechewelldefined] (3.2, 3.45) edge[out=50,in=190] (3.78, 3.7);	
		\draw[textlabel] (3.3, 3.76) node{\tiny{$\P(\emptyset)$}};
		
		\draw[cadrePof] (3.85, 3.48) rectangle (4.25, 3.765);		
		
		\draw[niveau] (1.5, 1) -- (5, 1);
		\draw[niveau] (1.5, 2) -- (5, 2);
		\draw[niveau] (1.5, 3) -- (5, 3);
		
		\draw[flecheordre] (1.2, 1) -- (1.2, 3);		
		
		\draw[textlabel] (0.8, 2) node{$\leq_\P$};
		
		\node[noeud] (1) at (2, 1) {};
		\node[noeudlabel, above of=1] {$p$};
		\node[noeud] (2) at (4.2, 1) {};
		\node[noeudlabel, above of=2] {$q$};
		\node[noeud] (3) at (3.1, 2) {};
		\node[noeudlabel, above of=3] {$pq$};
		\node[noeud] (4) at (3.1, 3) {};
		\node[noeudlabel, above of=4] {$\emptyset$};
	\end{tikzpicture}
	}
	\hspace*{.9cm}
	\subfigure[The SE models of $\Q_2$ are highlighted. We have $\SE(\Q_2) = \{(\emptyset, \emptyset), (pq, pq)\}$
	and $\SE(\P \star \Q_2) = \{(pq, pq)\}$.]{\label{subfig-GLPb}
		\begin{tikzpicture}[yscale=1.2,xscale=1]
		\tikzstyle{noeud}=[fill,circle,inner sep=1.5pt]
		\tikzstyle{niveau}=[-,line width=0.6pt]
		\tikzstyle{noeudlabel}=[node distance=0.3cm]
		\tikzstyle{flecheordre}=[->,dotted,line width=1pt,>=stealth]
		\tikzstyle{flechelabel}=[->,thin,>=stealth]
		\tikzstyle{textlabel}=[]
		\tikzstyle{cadrepsi}=[-, dashed, rounded corners=3pt]
		\tikzstyle{cadreresult}=[-, dashed, rounded corners=2pt, fill=gray!60, fill opacity=0.6]
		\tikzstyle{cadrewelldefined}=[-, very thin, rounded corners=3pt, dotted, fill=gray!60, fill opacity=0.15]
		\tikzstyle{linewelldefined}=[-, very thin]
		\tikzstyle{flechewelldefined}=[->,thin,>=stealth]
		\tikzstyle{cadrePof}=[-, rounded corners=2pt]
		
		
		\fill[cadreresult,transparent] (1.8, 0.85) rectangle (2.2, 1.4);
		\fill[cadreresult] (2.9, 2.85) rectangle (3.3, 3.4);
		\fill[cadreresult] (3.85, 3.48) rectangle (4.25, 3.765);
		\fill[cadreresult] (2.85, 1.85) rectangle (3.35, 2.4);
		\fill[cadreresult] (4.35, 2.12) rectangle (4.75, 2.45);
		
%
		
		\fill[cadrewelldefined] (2.7, 1.1) rectangle (3.8, 1.8);
		\draw[textlabel] (2.95, 1.63) node{$\emptyset$};
		\draw[textlabel] (3.45, 1.61) node{$p$};
		\draw[textlabel] (2.95, 1.26) node{$q$};
		\draw[textlabel] (3.45, 1.26) node{$pq$};
		\path[flechewelldefined] (2.1, 1.45) edge[out=50,in=190] (2.68, 1.7);	
		\draw[textlabel] (2.2, 1.76) node{\tiny{$\P(p)$}};
		
		\draw[cadrePof] (3.25, 1.48) rectangle (3.65, 1.75);
		
		\fill[cadrewelldefined] (4.9, 1.1) rectangle (6, 1.8);
		\draw[textlabel] (5.15, 1.63) node{$\emptyset$};
		\draw[textlabel] (5.65, 1.61) node{$p$};
		\draw[textlabel] (5.15, 1.26) node{$q$};
		\draw[textlabel] (5.65, 1.26) node{$pq$};
		\path[flechewelldefined] (4.3, 1.45) edge[out=50,in=190] (4.88, 1.7);	
		\draw[textlabel] (4.4, 1.76) node{\tiny{$\P(q)$}};
		
		\draw[cadrePof] (4.95, 1.12) rectangle (5.35, 1.765);
		
		\fill[cadrewelldefined] (3.8, 2.1) rectangle (4.9, 2.8);
		\draw[textlabel] (4.05, 2.63) node{$\emptyset$};
		\draw[textlabel] (4.55, 2.61) node{$p$};
		\draw[textlabel] (4.05, 2.26) node{$q$};
		\draw[textlabel] (4.55, 2.26) node{$pq$};
		\path[flechewelldefined] (3.2, 2.45) edge[out=50,in=190] (3.78, 2.7);	
		\draw[textlabel] (3.3, 2.76) node{\tiny{$\P(pq)$}};
		
		\draw[cadrePof] (4.35, 2.12) rectangle (4.75, 2.75);
		
		\fill[cadrewelldefined] (3.8, 3.1) rectangle (4.9, 3.8);
		\draw[textlabel] (4.05, 3.63) node{$\emptyset$};
		\draw[textlabel] (4.55, 3.61) node{$p$};
		\draw[textlabel] (4.05, 3.26) node{$q$};
		\draw[textlabel] (4.55, 3.26) node{$pq$};
		\path[flechewelldefined] (3.2, 3.45) edge[out=50,in=190] (3.78, 3.7);	
		\draw[textlabel] (3.3, 3.76) node{\tiny{$\P(\emptyset)$}};
		
		\draw[cadrePof] (3.85, 3.48) rectangle (4.25, 3.765);
		
		\draw[niveau] (1.5, 1) -- (5, 1);
		\draw[niveau] (1.5, 2) -- (5, 2);
		\draw[niveau] (1.5, 3) -- (5, 3);
		
		\draw[flecheordre] (1.2, 1) -- (1.2, 3);		
		
		\draw[textlabel] (0.8, 2) node{$\leq_\P$};
		
		\node[noeud] (1) at (2, 1) {};
		\node[noeudlabel, above of=1] {$p$};
		\node[noeud] (2) at (4.2, 1) {};
		\node[noeudlabel, above of=2] {$q$};
		\node[noeud] (3) at (3.1, 2) {};
		\node[noeudlabel, above of=3] {$pq$};
		\node[noeud] (4) at (3.1, 3) {};
		\node[noeudlabel, above of=4] {$\emptyset$};
	\end{tikzpicture}
	}
\caption{The SE models of $\Q_1$ and $\Q_2$ highlighted within $\leq_\P$ and sets $\P(Y)$ for each interpretation $Y$.}
\label{fig:GLP-parted-assignment2}
\end{figure}
\label{ex:first-GLP-revision}
\end{example}

Due to the similarities between an LP faithful assignment (cf. Definition \ref{def:LPfaithful}) and a faithful assignment
(cf. Definition \ref{def:faithful}), an interesting consequence from
Theorem \ref{proposition : AGM revision operator caracterization} and Proposition
\ref{prop:characterization-revision-GLP} is that
%
every GLP revision operator can be viewed as an extension of a (propositional) KM revision operator:

\begin{definition}[Propositional-based LP revision operator]
Let $\circ$ be a propositional revision operator
and $f$ be a mapping from $\allinterpretations$ to $2^\allinterpretations$ such that
for every interpretation $Y$, $Y \in f(Y)$ and if $X \in f(Y)$ then $X \subseteq Y$.
The \emph{propositional-based LP revision operator} w.r.t. $\circ$ and $f$,
denoted $\star^{\circ,f}$, is defined for all GLPs $\P, \Q$ by
$$\SE(\P \star^{\circ,f} \Q) =\left\lbrace\begin{array}{ll}
\SE(\P + \Q) \hfill \mbox{ if $\P + \Q$ is consistent,}\\
\{(X, Y) \in \SE(\Q) \mid Y \models \formGLPb{\P} \circ \formGLPb{\Q}, X \in f(Y)\} \hfill \mbox{ \ \ \ \ \ \ \ \ \ \ otherwise.}
\end{array}\right.$$

$\star^{\circ,f}$ is said to be a \emph{propositional-based GLP revision operator} if $\circ$
is a KM revision operator (i.e., satisfying postulates (R1 - R6)).
\label{def:propbasedGLPoperator}
\end{definition}



\begin{proposition}
An LP revision operator is a GLP revision operator if and only if it is a propositional-based GLP revision operator.
\label{prop:prop-based equl GLP}
\end{proposition}

\nico{
In the previous section, we noticed that there is a one-to-one correspondence between the
KM revision operators (modulo equivalence) and the set of all faithful assignments (cf. Proposition \ref{rem:faithful}).
Interestingly, we get a similar result
in the case of GLP revision operators with respect to propositional-based GLP revision operators
(cf. Corollary \ref{cor:cor GLP correspondence} below). Let us introduce an intermediate result:
\begin{proposition}
For all propositional-based GLP revision operators $\star^{\circ_1, f_1}, \star^{\circ_2, f_2}$,
we have $\star^{\circ_1, f_1} = \star^{\circ_2, f_2}$ if and only if $\circ_1 = \circ_2$ and $f_1 = f_2$.
\label{prop:prop-GLP unique}
\end{proposition}
This proposition tells us that if $\circ_1 \neq \circ_2$
or $f_1 \neq f_2$, then for some pair of GLPs $\P, \Q$ we will get $\P \star^{\circ_1, f_1} \Q \not\equiv \P \star^{\circ_2, f_2} \Q$,
that is to say, different choices of parameters for a propositional-based LP revision operator lead to different propositional-based LP revision operators.
As a direct consequence of Propositions \ref{prop:prop-based equl GLP} and \ref{prop:prop-GLP unique}, we get that:
\begin{corollary}
There is a one-to-one correspondence between the set of GLP revision operators and the set of propositional-based
GLP revision operators.
\label{cor:cor GLP correspondence}
\end{corollary}
}



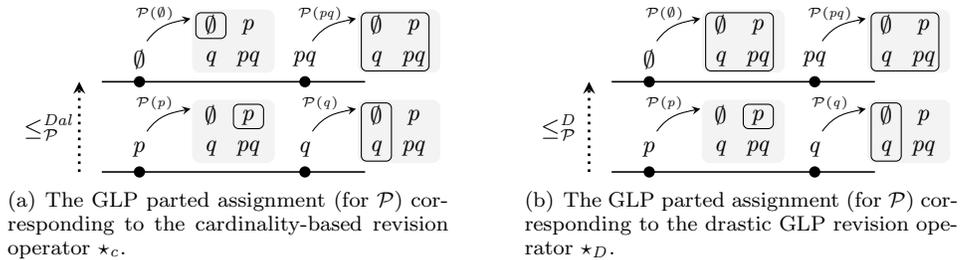
\begin{figure}[!htb]
	\centering
	\subfigure[The GLP parted assignment (for $\P$) corresponding to the cardinality-based revision operator $\star_c$.]{\label{subfig-GLP-card-drastic1}
		\begin{tikzpicture}[yscale=1.2,xscale=1]
		\tikzstyle{noeud}=[fill,circle,inner sep=1.5pt]
		\tikzstyle{niveau}=[-,line width=0.6pt]
		\tikzstyle{noeudlabel}=[node distance=0.3cm]
		\tikzstyle{flecheordre}=[->,dotted,line width=1pt,>=stealth]
		\tikzstyle{flechelabel}=[->,thin,>=stealth]
		\tikzstyle{textlabel}=[]
		\tikzstyle{cadrepsi}=[-, dashed, rounded corners=3pt]
		\tikzstyle{cadreresult}=[-, dashed, rounded corners=2pt, fill=gray!60, fill opacity=0.6]
		\tikzstyle{cadrewelldefined}=[-, very thin, rounded corners=3pt, dotted, fill=gray!60, fill opacity=0.15]
		\tikzstyle{linewelldefined}=[-, very thin]
		\tikzstyle{flechewelldefined}=[->,thin,>=stealth]
		\tikzstyle{cadrePof}=[-, rounded corners=2pt]
		
		
		
%
		
		\fill[cadrewelldefined] (2.7, 1.1) rectangle (3.8, 1.8);
		\draw[textlabel] (2.95, 1.63) node{$\emptyset$};
		\draw[textlabel] (3.45, 1.61) node{$p$};
		\draw[textlabel] (2.95, 1.26) node{$q$};
		\draw[textlabel] (3.45, 1.26) node{$pq$};
		\path[flechewelldefined] (2.1, 1.45) edge[out=50,in=190] (2.68, 1.7);
		\draw[textlabel] (2.2, 1.76) node{\tiny{$\P(p)$}};
		
		\draw[cadrePof] (3.25, 1.48) rectangle (3.65, 1.75);
		
		\fill[cadrewelldefined] (4.9, 1.1) rectangle (6, 1.8);
		\draw[textlabel] (5.15, 1.63) node{$\emptyset$};
		\draw[textlabel] (5.65, 1.61) node{$p$};
		\draw[textlabel] (5.15, 1.26) node{$q$};
		\draw[textlabel] (5.65, 1.26) node{$pq$};
		\path[flechewelldefined] (4.3, 1.45) edge[out=50,in=190] (4.88, 1.7);	
		\draw[textlabel] (4.4, 1.76) node{\tiny{$\P(q)$}};
		
		\draw[cadrePof] (4.95, 1.12) rectangle (5.35, 1.765);
		
		\fill[cadrewelldefined] (4.9, 2.1) rectangle (6, 2.8);
		\draw[textlabel] (5.15, 2.63) node{$\emptyset$};
		\draw[textlabel] (5.65, 2.61) node{$p$};
		\draw[textlabel] (5.15, 2.26) node{$q$};
		\draw[textlabel] (5.65, 2.26) node{$pq$};
		\path[flechewelldefined] (4.3, 2.45) edge[out=50,in=190] (4.88, 2.7);	
		\draw[textlabel] (4.4, 2.76) node{\tiny{$\P(pq)$}};
		
		\draw[cadrePof] (4.95, 2.12) rectangle (5.85, 2.765);
		
		\fill[cadrewelldefined] (2.7, 2.1) rectangle (3.8, 2.8);
		\draw[textlabel] (2.95, 2.63) node{$\emptyset$};
		\draw[textlabel] (3.45, 2.61) node{$p$};
		\draw[textlabel] (2.95, 2.26) node{$q$};
		\draw[textlabel] (3.45, 2.26) node{$pq$};
		\path[flechewelldefined] (2.1, 2.45) edge[out=50,in=190] (2.68, 2.7);	
		\draw[textlabel] (2.2, 2.76) node{\tiny{$\P(\emptyset)$}};
		
		\draw[cadrePof] (2.75, 2.48) rectangle (3.15, 2.765);
		
		\draw[niveau] (1.5, 1) -- (5, 1);
		\draw[niveau] (1.5, 2) -- (5, 2);
		
		\draw[flecheordre] (1.2, 1) -- (1.2, 2);		
		
		\draw[textlabel] (0.8, 1.5) node{$\leq^{Dal}_\P$};
		
		\node[noeud] (1) at (2, 1) {};
		\node[noeudlabel, above of=1] {$p$};
		\node[noeud] (2) at (4.2, 1) {};
		\node[noeudlabel, above of=2] {$q$};
		\node[noeud] (3) at (4.2, 2) {};
		\node[noeudlabel, above of=3] {$pq$};
		\node[noeud] (4) at (2, 2) {};
		\node[noeudlabel, above of=4] {$\emptyset$};
	\end{tikzpicture}
	}
	\hspace*{.8cm}
	\subfigure[The GLP parted assignment (for $\P$) corresponding to the drastic GLP revision operator $\star_D$.]{\label{subfig-GLP-card-drastic2}
		\begin{tikzpicture}[yscale=1.2,xscale=1]
		\tikzstyle{noeud}=[fill,circle,inner sep=1.5pt]
		\tikzstyle{niveau}=[-,line width=0.6pt]
		\tikzstyle{noeudlabel}=[node distance=0.3cm]
		\tikzstyle{flecheordre}=[->,dotted,line width=1pt,>=stealth]
		\tikzstyle{flechelabel}=[->,thin,>=stealth]
		\tikzstyle{textlabel}=[]
		\tikzstyle{cadrepsi}=[-, dashed, rounded corners=3pt]
		\tikzstyle{cadreresult}=[-, dashed, rounded corners=2pt, fill=gray!60, fill opacity=0.6]
		\tikzstyle{cadrewelldefined}=[-, very thin, rounded corners=3pt, dotted, fill=gray!60, fill opacity=0.15]
		\tikzstyle{linewelldefined}=[-, very thin]
		\tikzstyle{flechewelldefined}=[->,thin,>=stealth]
		\tikzstyle{cadrePof}=[-, rounded corners=2pt]
		
		
		
%
		
		\fill[cadrewelldefined] (2.7, 1.1) rectangle (3.8, 1.8);
		\draw[textlabel] (2.95, 1.63) node{$\emptyset$};
		\draw[textlabel] (3.45, 1.61) node{$p$};
		\draw[textlabel] (2.95, 1.26) node{$q$};
		\draw[textlabel] (3.45, 1.26) node{$pq$};
		\path[flechewelldefined] (2.1, 1.45) edge[out=50,in=190] (2.68, 1.7);
		\draw[textlabel] (2.2, 1.76) node{\tiny{$\P(p)$}};
		
		\draw[cadrePof] (3.25, 1.48) rectangle (3.65, 1.75);
		
		\fill[cadrewelldefined] (4.9, 1.1) rectangle (6, 1.8);
		\draw[textlabel] (5.15, 1.63) node{$\emptyset$};
		\draw[textlabel] (5.65, 1.61) node{$p$};
		\draw[textlabel] (5.15, 1.26) node{$q$};
		\draw[textlabel] (5.65, 1.26) node{$pq$};
		\path[flechewelldefined] (4.3, 1.45) edge[out=50,in=190] (4.88, 1.7);	
		\draw[textlabel] (4.4, 1.76) node{\tiny{$\P(q)$}};
		
		\draw[cadrePof] (4.95, 1.12) rectangle (5.35, 1.765);
		
		\fill[cadrewelldefined] (4.9, 2.1) rectangle (6, 2.8);
		\draw[textlabel] (5.15, 2.63) node{$\emptyset$};
		\draw[textlabel] (5.65, 2.61) node{$p$};
		\draw[textlabel] (5.15, 2.26) node{$q$};
		\draw[textlabel] (5.65, 2.26) node{$pq$};
		\path[flechewelldefined] (4.3, 2.45) edge[out=50,in=190] (4.88, 2.7);	
		\draw[textlabel] (4.4, 2.76) node{\tiny{$\P(pq)$}};
		
		\draw[cadrePof] (4.95, 2.12) rectangle (5.85, 2.765);
		
		\fill[cadrewelldefined] (2.7, 2.1) rectangle (3.8, 2.8);
		\draw[textlabel] (2.95, 2.63) node{$\emptyset$};
		\draw[textlabel] (3.45, 2.61) node{$p$};
		\draw[textlabel] (2.95, 2.26) node{$q$};
		\draw[textlabel] (3.45, 2.26) node{$pq$};
		\path[flechewelldefined] (2.1, 2.45) edge[out=50,in=190] (2.68, 2.7);	
		\draw[textlabel] (2.2, 2.76) node{\tiny{$\P(\emptyset)$}};
		
		\draw[cadrePof] (2.75, 2.12) rectangle (3.65, 2.765);
		
		\draw[niveau] (1.5, 1) -- (5, 1);
		\draw[niveau] (1.5, 2) -- (5, 2);
		
		\draw[flecheordre] (1.2, 1) -- (1.2, 2);		
		
		\draw[textlabel] (0.8, 1.5) node{$\leq^{D}_\P$};
		
		\node[noeud] (1) at (2, 1) {};
		\node[noeudlabel, above of=1] {$p$};
		\node[noeud] (2) at (4.2, 1) {};
		\node[noeudlabel, above of=2] {$q$};
		\node[noeud] (3) at (4.2, 2) {};
		\node[noeudlabel, above of=3] {$pq$};
		\node[noeud] (4) at (2, 2) {};
		\node[noeudlabel, above of=4] {$\emptyset$};
	\end{tikzpicture}
	}
\caption{The GLP parted assignments corresponding to the cardinality-based and drastic GLP revision operators, focusing on the GLP $\P$.}
\label{fig:GLP-parted-assignment3}
\end{figure}

Note that the cardinality-based revision operator $\star_c$ (cf. Definition \ref{def:cardinal-based}) corresponds
to the propositional-based GLP revision operator $\star^{\circ_{Dal}, f_1}$, where $\circ_{Dal}$ is the
Dalal revision operator
(cf. Definition \ref{def:dalal})
and $f_1$ is defined for every interpretation $Y$ as
$f_1(Y) = \{X \in \Omega \mid X \subseteq Y$ and if $X \subsetneq Y$ then
$X \models \alpha_{(\P, Y)} \circ_{Dal} \alpha_Y\}$, where $\alpha_Y$ is any propositional
formula such that $mod(\alpha_Y) = \{X \in \Omega \mid X \subseteq Y\}$,
$\alpha_{(\P, Y)}$ is any propositional formula satisfying
$mod(\alpha_{(\P, Y)}) = \{X \in \Omega \mid (X, Y') \in \SE(\P), Y' \models form_Y \circ_{Dal} \alpha^2_\P\}$,
and $form_Y$ is any propositional formula satisfying $mod(form_Y) = \{Y\}$.
In addition, the drastic GLP revision operator (cf. Definition \nolinebreak\ref{def:drastic}) corresponds
to the propositional-based GLP revision operator $\star^{\circ_D, f_2}$, where $\circ_D$ is the
drastic revision operator (cf. Definition \ref{def:drasticr}) and $f_2$ is defined for every interpretation $Y$ as
$f_2(Y) = 2^Y$. Figures \ref{subfig-GLP-card-drastic1} and \ref{subfig-GLP-card-drastic2} provide the
graphical representation of these two operators in terms of parted assignments similarly to Figure \ref{fig:GLP-parted-assignment1},
focusing on the GLP $\P$ from Example \ref{ex:detailed GLP P}.

Remark that in the case where $\P$ and $\Q$ have no common SE models, then
a (propositional-based) GLP revision operator $\star^{\circ, f}$
``rejects'' as candidates for the SE models of the revised program $\P \star^{\circ, f} \Q$
those SE interpretations whose second component is not a classical model of $\formGLPb{\P} \circ \formGLPb{\Q}$;
that is to say, as an upstream selection step the potential resulting SE models are chosen
with respect to their second component
by the underlying propositional revision operator $\circ$.
Then, one can see from Definition \ref{def:propbasedGLPoperator}
that the function $f$ is used as a second filtering step that is made with respect to the first component of those preselected SE interpretations,
and that this final selection becomes independent of the underlying input program $\P$.
Then it becomes questionable whether the postulates (RA1 - RA6) sufficiently describe the rational behavior
of LP revision operators.
Indeed, we will show in the next section that this ``freedom'' on the definition of the function $f$
raises some issues for some specific subclasses of fully rational LP revision operators.

\subsection{Comparison with other existing works}
As we already briefly mentionned in the introduction, Delgrande \textit{et al.\/} \citeyear{Delgrande/LPNMR} also recently proposed a
constructive characterization of belief revision operators for logic programs that satisfy the whole set of postulates (RA1 - RA6).
They considered various forms of logic programs, i.e., generalized, disjunctive, normal, positive, and Horn, so we shall now compare
our characterization with the one given in \cite{Delgrande/LPNMR} for the case of GLPs:
\begin{definition}[GLP compliant faithful assignment \cite{Delgrande/LPNMR}]
A \emph{GLP compliant faithful assignment} is a mapping which associates every GLP $\P$ with a total preorder
$\leq^*_\P$ over SE interpretations such that for all GLPs $\P, \Q$ and all SE interpretations
$(X, Y), (X', Y')$,
the following conditions hold:
\begin{description}
\item[(1)] If $(X, Y) \in \SE(\P)$ and $(X', Y') \in \SE(\P)$, then $(X, Y) \simeq^*_\P (X', Y')$;
\item[(2)] If $(X, Y) \in \SE(\P)$ and $(X', Y') \not\in \SE(\P)$, then $(X, Y) <^*_\P (X', Y')$;
\item[(3)] If $\P \equiv_s \Q$, then $\leq_\P = \leq_\Q$;
\item[(4)] $(Y, Y) \leq^*_\P (X, Y)$.
\end{description}
\label{def:delgrande-compliant}
\end{definition}

The following theorem is expressed as a combination
of Theorems 4 and 5 from \cite{Delgrande/LPNMR} applied to GLPs:

\begin{theorem}[Delgrande et al. 2013a]
An LP revision operator $\star$ is a GLP revision operator (i.e., it satisfies postulates (RA1 - RA6)) if and only if
there exists a GLP compliant faithful assignment associating every GLP $\P$ with a total preorder $\leq^*_\P$ such that
for all GLPs $\P, \Q$, $\SE(\P \star \Q) = \min(\SE(\Q), \leq^*_\P)$.
\footnote{In \cite{Delgrande/LPNMR}, an additional postulate is considered in the characterization theorems, namely (Acyc). However,
it is harmless to omit this postulate here since (Acyc) is a logical consequence of the postulates (RA1 - RA6) in the case
of generalized logic programs (cf. \cite{Delgrande/LPNMR}, Theorem 2).}
\label{th:delgrande-charac}
\end{theorem}

Since both our GLP parted assignments and Delgrande \textit{et al.\/}'s GLP compliant faithful assignments characterize the class of
GLP revision operators, there must exist a relationship between the two structures.
We denote by $GLP_{part}$ the set of all GLP parted assignments and $GLP_{faith}$ the set of all
GLP compliant faithful assignments. We now formally establish a correspondence between the two sets.

\begin{definition}
Let $\sigma_{part\rightarrow faith}$ be a binary relation on $GLP_{part} \times GLP_{faith}$ defined as follows.
For every $(\Phi, \Psi) \in GLP_{part}$ (where $\Phi$ associates every GLP $\P$ with a total preorder $\leq_\P$, and
$\Psi$ associates every GLP $\P$ and every interpretation $Y$ with a set of interpretations $\P(Y)$), and for every
$\Gamma \in GLP_{faith}$ (where $\Gamma$ associates every GLP $\P$ with a total preorder $\leq^*_\P$),
we have $((\Phi, \Psi), \Gamma) \in \sigma_{part\rightarrow faith}$ if and only if for every GLP $\P$, for all interpretations
$X, Y, Y'$, $X \subseteq Y$,
the following conditions are satisfied:
\begin{itemize}
\item[(i)] $(Y, Y) \leq^*_\P (Y', Y')$ if and only if $Y \leq_\P Y'$, and
\item[(ii)] $(X, Y) \leq^*_\P (Y, Y)$ if and only if $X \in \P(Y)$.
\end{itemize}
\label{def:mapping-between-two-structures}
\end{definition}

We show now that a pair of assignments from $GLP_{part} \times GLP_{faith}$ satisfies the relation $\sigma_{part\rightarrow faith}$
if and only if represent both assignments represent the same GLP revision operator:

\begin{proposition}
For every $(\Phi, \Psi) \in GLP_{part}$ and every $\Gamma \in GLP_{faith}$, $((\Phi, \Psi), \Gamma) \in \sigma_{part\rightarrow faith}$
if and only if for all GLPs $\P, \Q$,
$\min(\SE(\Q), \leq^*_\P) = \{(X, Y) \mid (X, Y) \in \SE(\Q),
Y \in \min(\modeles{\Q},$ $\leq_\P),
X \in \P(Y)\}$.
\label{prop:sigma-injective}
\end{proposition}

Whereas our GLP parted assignments are formed of two structures which are independent from each other
(an LP faithful assignment used to order the second components of SE interpretations, and a well-defined assignment
selecting the first component of SE interpretations),
Delgrande \textit{et al.\/}'s GLP compliant faithful assignments consist of a single structure, i.e.,
a set of total preoders over SE interpretations.
Though it may look simpler to represent a GLP revision operator through a single assignment, it turns out that
the induced characterization (cf. Theorem \ref{th:delgrande-charac}) is not a one-to-one correspondence;
more precisely, $\sigma_{part\rightarrow faith}$ is not a function
and as a consequence, a given GLP revision operator can be represented by different GLP compliant faithful assignments.
Roughly speaking, this is due to the fact that totality required by preorders $\leq^*_\P$
is actually not needed. Many comparisons between pairs of SE interpretations within a total preorder $\leq^*_\P$
are irrelevant to the GLP revision operator they correspond to. This is illustrated in the following example:

\begin{example}
Consider again the GLP $\P$ from Example \ref{ex:detailed GLP P} and the GLP parted assignment $(\Phi, \Psi)$ focusing on $\P$
depicted in Figure \ref{fig:GLP-parted-assignment1}. Then Figure \ref{fig:3-GLP-compliant-faithful-assignments} depicts three total preorders $\leq^1_\P$,
$\leq^2_\P$ and $\leq^3_\P$ induced from three different GLP compliant faithful assignments $\Gamma^1$, $\Gamma^2$ and $\Gamma^3$
which both correspond to the GLP parted assignment $(\Phi, \Psi)$, i.e.,
$((\Phi, \Psi), \Gamma^1), ((\Phi, \Psi), \Gamma^2), ((\Phi, \Psi), \Gamma^3) \in \sigma_{part \rightarrow faith}$.
It can be easily
checked
that for any GLP $\Q$, $\min(\SE(\Q), \leq^1_\P) = \min(\SE(\Q), \leq^2_\P) = \min(\SE(\Q), \leq^3_\P)$.
The SE interpretations enclosed in dashed boxes correspond to those $(X, Y) \in \{(\emptyset, p), (\emptyset, pq), (q, pq)\}$
whose comparison with other SE interpretations is irrelevant to the represented GLP revision operator, as far as one has $(Y, Y) <^i_\P (X, Y)$ for
$i \in \{1, 2, 3\}$.
\begin{figure}[!htb]
	\centering
	\subfigure[The total preorder $\leq^1_\P$ associated with $\Gamma^1$.]{\label{subfig-GLP-compliant-faithful1}
		\begin{tikzpicture}[yscale=1.5,xscale=1]
		\tikzstyle{noeud}=[fill,circle,inner sep=1.5pt]
		\tikzstyle{niveau}=[-,line width=0.6pt]
		\tikzstyle{noeudlabel}=[node distance=0.3cm]
		\tikzstyle{flecheordre}=[->,dotted,line width=1pt,>=stealth]
		\tikzstyle{flechelabel}=[->,thin,>=stealth]
		\tikzstyle{textlabel}=[]
		\tikzstyle{cadrepsi}=[-, dashed, rounded corners=3pt]
		\tikzstyle{cadreresult}=[-, dashed, rounded corners=2pt, fill=gray!60, fill opacity=0.6]
		\tikzstyle{cadrewelldefined}=[-, very thin, rounded corners=3pt, dotted, fill=gray!60, fill opacity=0.15]
		\tikzstyle{linewelldefined}=[-, very thin]
		\tikzstyle{flechewelldefined}=[->,thin,>=stealth]
		\tikzstyle{cadrePof}=[-, rounded corners=2pt, dashed]
		
		
		
		\draw[cadrePof] (1.55, 1.85) rectangle (2.45, 2.4);
		\draw[cadrePof] (1.5, 2.85) rectangle (3.5, 3.4);
		
		\draw[niveau] (1.6, 1) -- (4.4, 1);
		\draw[niveau] (1.6, 2) -- (4.4, 2);
		\draw[niveau] (1.6, 3) -- (4.4, 3);
		
		\draw[flecheordre] (1.4, 1) -- (1.4, 3);		
		
		\draw[textlabel] (1.1, 2) node{$\leq^1_\P$};
		
		\node[noeud] (1) at (2, 1) {};
		\node[noeudlabel, above of=1] {$(p, p)$};
		\node[noeud] (2) at (3, 1) {};
		\node[noeudlabel, above of=2] {$(\emptyset, q)$};
		\node[noeud] (3) at (4, 1) {};
		\node[noeudlabel, above of=3] {$(q, q)$};
		\node[noeud] (4) at (2, 2) {};
		\node[noeudlabel, above of=4] {$(\emptyset, p)$};
		\node[noeud] (5) at (3, 2) {};
		\node[noeudlabel, above of=5] {$(p, pq)$};
		\node[noeud] (6) at (4, 2) {};
		\node[noeudlabel, above of=6] {$(pq, pq)$};
		\node[noeud] (7) at (2, 3) {};
		\node[noeudlabel, above of=7] {$(\emptyset, pq)$};
		\node[noeud] (8) at (3, 3) {};
		\node[noeudlabel, above of=8] {$(q, pq)$};
		\node[noeud] (9) at (4, 3) {};
		\node[noeudlabel, above of=9] {$(\emptyset, \emptyset)$};
	\end{tikzpicture}
	}
	\subfigure[The total preorder $\leq^2_\P$ associated with $\Gamma^2$.]{\label{subfig-GLP-compliant-faithful2}
	\begin{tikzpicture}[yscale=1.5,xscale=1]
		\tikzstyle{noeud}=[fill,circle,inner sep=1.5pt]
		\tikzstyle{niveau}=[-,line width=0.6pt]
		\tikzstyle{noeudlabel}=[node distance=0.3cm]
		\tikzstyle{flecheordre}=[->,dotted,line width=1pt,>=stealth]
		\tikzstyle{flechelabel}=[->,thin,>=stealth]
		\tikzstyle{textlabel}=[]
		\tikzstyle{cadrepsi}=[-, dashed, rounded corners=3pt]
		\tikzstyle{cadreresult}=[-, dashed, rounded corners=2pt, fill=gray!60, fill opacity=0.6]
		\tikzstyle{cadrewelldefined}=[-, very thin, rounded corners=3pt, dotted, fill=gray!60, fill opacity=0.15]
		\tikzstyle{linewelldefined}=[-, very thin]
		\tikzstyle{flechewelldefined}=[->,thin,>=stealth]
		\tikzstyle{cadrePof}=[-, rounded corners=2pt, dashed]
		
		
		
		\draw[cadrePof] (2, 2.18) rectangle (3, 2.7);
		\draw[cadrePof] (2.05, 2.85) rectangle (4, 3.4);
		
		\draw[niveau] (1.6, 1) -- (4.4, 1);
		\draw[niveau] (1.6, 1.67) -- (4.4, 1.67);
		\draw[niveau] (1.6, 2.33) -- (4.4, 2.33);
		\draw[niveau] (1.6, 3) -- (4.4, 3);
		
		\draw[flecheordre] (1.4, 1) -- (1.4, 3);		
		
		\draw[textlabel] (1.1, 2) node{$\leq^2_\P$};
		
		\node[noeud] (1) at (2, 1) {};
		\node[noeudlabel, above of=1] {$(p, p)$};
		\node[noeud] (2) at (3, 1) {};
		\node[noeudlabel, above of=2] {$(\emptyset, q)$};
		\node[noeud] (3) at (4, 1) {};
		\node[noeudlabel, above of=3] {$(q, q)$};
		\node[noeud] (4) at (2.5, 2.33) {};
		\node[noeudlabel, above of=4] {$(\emptyset, pq)$};
		\node[noeud] (5) at (2.5, 1.67) {};
		\node[noeudlabel, above of=5] {$(p, pq)$};
		\node[noeud] (6) at (3.5, 1.67) {};
		\node[noeudlabel, above of=6] {$(pq, pq)$};
		\node[noeud] (7) at (2.5, 3) {};
		\node[noeudlabel, above of=7] {$(\emptyset, p)$};
		\node[noeud] (8) at (3.5, 3) {};
		\node[noeudlabel, above of=8] {$(q, pq)$};
		\node[noeud] (9) at (3.5, 2.33) {};
		\node[noeudlabel, above of=9] {$(\emptyset, \emptyset)$};
	\end{tikzpicture}
	}
	\subfigure[The total preorder $\leq^3_\P$ associated with $\Gamma^3$.]{\label{subfig-GLP-compliant-faithful3}
	\begin{tikzpicture}[yscale=1.5,xscale=1]
		\tikzstyle{noeud}=[fill,circle,inner sep=1.5pt]
		\tikzstyle{niveau}=[-,line width=0.6pt]
		\tikzstyle{noeudlabel}=[node distance=0.3cm]
		\tikzstyle{flecheordre}=[->,dotted,line width=1pt,>=stealth]
		\tikzstyle{flechelabel}=[->,thin,>=stealth]
		\tikzstyle{textlabel}=[]
		\tikzstyle{cadrepsi}=[-, dashed, rounded corners=3pt]
		\tikzstyle{cadreresult}=[-, dashed, rounded corners=2pt, fill=gray!60, fill opacity=0.6]
		\tikzstyle{cadrewelldefined}=[-, very thin, rounded corners=3pt, dotted, fill=gray!60, fill opacity=0.15]
		\tikzstyle{linewelldefined}=[-, very thin]
		\tikzstyle{flechewelldefined}=[->,thin,>=stealth]
		\tikzstyle{cadrePof}=[-, rounded corners=2pt, dashed]
		
		
		
		\draw[cadrePof] (2.55, 1.38) rectangle (3.45, 1.87);
		\draw[cadrePof] (2, 2.38) rectangle (3, 2.87);
		\draw[cadrePof] (2.5, 2.9) rectangle (3.5, 3.37);
		
		\draw[niveau] (1.6, 1) -- (4.4, 1);
		\draw[niveau] (1.6, 1.5) -- (4.4, 1.5);
		\draw[niveau] (1.6, 2) -- (4.4, 2);
		\draw[niveau] (1.6, 2.5) -- (4.4, 2.5);
		\draw[niveau] (1.6, 3) -- (4.4, 3);
		
		\draw[flecheordre] (1.4, 1) -- (1.4, 3);		
		
		\draw[textlabel] (1.1, 2) node{$\leq^3_\P$};
		
		\node[noeud] (1) at (2, 1) {};
		\node[noeudlabel, above of=1] {$(p, p)$};
		\node[noeud] (2) at (3, 1) {};
		\node[noeudlabel, above of=2] {$(\emptyset, q)$};
		\node[noeud] (3) at (4, 1) {};
		\node[noeudlabel, above of=3] {$(q, q)$};
		\node[noeud] (4) at (3, 1.5) {};
		\node[noeudlabel, above of=4] {$(\emptyset, p)$};
		\node[noeud] (5) at (2.5, 2) {};
		\node[noeudlabel, above of=5] {$(p, pq)$};
		\node[noeud] (6) at (3.5, 2) {};
		\node[noeudlabel, above of=6] {$(pq, pq)$};
		\node[noeud] (7) at (3, 3) {};
		\node[noeudlabel, above of=7] {$(\emptyset, pq)$};
		\node[noeud] (8) at (2.5, 2.5) {};
		\node[noeudlabel, above of=8] {$(q, pq)$};
		\node[noeud] (9) at (3.5, 2.5) {};
		\node[noeudlabel, above of=9] {$(\emptyset, \emptyset)$};
	\end{tikzpicture}
	}
\caption{Three total preorders corresponding to three different GLP compliant faithful assignments which correspond to the same GLP parted assignment.}
\label{fig:3-GLP-compliant-faithful-assignments}
\end{figure}
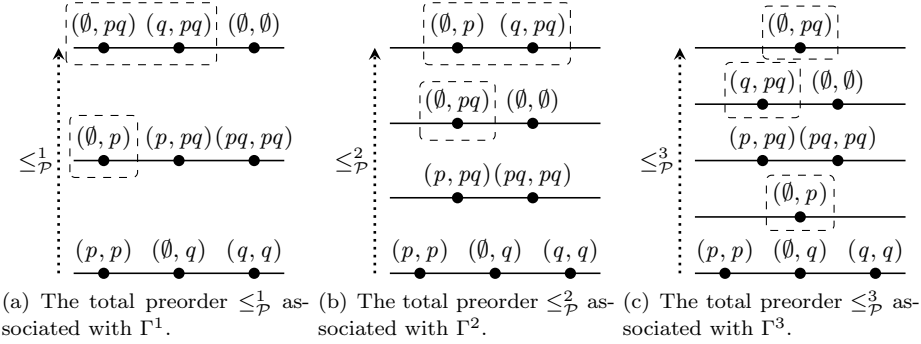
\end{example}

In fact, one can see that as soon as the language contains at least two propositional variables, e.g., $\{p, q\} \subseteq \mathcal{A}$ with $p \neq q$,
then the GLP $\P$ satisfying $(p, p), (q, q)$ $\in \SE(\P)$ and $(\emptyset, p), (\emptyset, q) \notin \SE(\P)$ can be
associated through a GLP compliant faithful assignment with at least three different total preorders;
an arbitrary relative ordering between the SE interpretations $(\emptyset, p)$ and $(\emptyset, q)$ will have no effect on the corresponding GLP revision operator.

Removing the property of totality from preorders involved in a GLP compliant faithful could be an alternative towards establishing
another one-to-one correspondence with GLP revision operators. However, our GLP parted assignments make clear the different roles played
by the first and second components of SE interpretations in terms of GLP revision. One the one hand the second components are totally ordered,
on the other hand the first components are arbitrarily selected as possible condidates for SE interpretations. This allows us to make precise the link with
propositional faithful assignments and propositional revision operators, which would not be clear with a slight adjustment of GLP compliant faithful assignments.
The next section shows how our propositional-based GLP revision operator facilitate the comprehension and analysis of GLP revision.

\section{GLP revision operators embedded into Boolean lattices}
\label{sec:lattice}

%

For every propositional revision operator $\circ$, let $\GLPops{\circ}$ denote
the set of all \linebreak propositional-based LP revision operators w.r.t. $\circ$.
One can remark that from Proposition \ref{prop:prop-GLP unique},
the set $\{\GLPops{\circ} \mid \circ \mbox{ is a KM revision operator}\}$ forms
a partition of the class of all GLP revision operators.
Let us now take a closer look to the set of GLP revision operators $\GLPops{\circ}$
when we are given any specific KM revision operator $\circ$:


\begin{definition}
Let $\circ$ be a propositional revision operator.
We define the binary relation $\lattice{\circ}$
over $\GLPops{\circ}$ as follows:
for all propositional-based LP revision operators $\star^{\circ,f_1}, \star^{\circ,f_2}$,
$\star^{\circ,f_1} \lattice{\circ} \star^{\circ,f_2}$ if and only for every interpretation $Y$,
we have $f_2(Y) \subseteq f_1(Y)$.
\label{def:relation GLPs}
\end{definition}

One can see that for each revision operator $\circ$, the set $(\GLPops{\circ}, \lattice{\circ})$
forms a structure that is isomorphic to a Boolean lattice\footnote{A Boolean lattice is a partially ordered set
$(E, \leq_E)$ which is isomorphic to the
set of subsets of some set $F$ together with the usual set-inclusion operation,
i.e., $(2^F, \subseteq)$.}, and the careful reader will notice that $(\GLPops{\circ}, \lattice{\circ})$
precisely corresponds to the product of the Boolean lattices
$\{(\mathbb{B}_Y, \subseteq) \mid Y \in \allinterpretations\}$, where
$\mathbb{B}_Y = \{Z \cup \{Y\} \mid Z \in 2^{2^Y \setminus Y}\}$.
The following result shows that this lattice structure can be used to
analyse the relative semantic behavior of GLP revision operators from $(\GLPops{\circ}, \lattice{\circ})$.

\begin{proposition}
Let $\circ$ be a KM revision operator.
It holds that for all GLP revision operators $\star_1, \star_2 \in \GLPops{\circ}$,
$\star_1 \lattice{\circ} \star_2$ if and only if for all GLPs $\P, \Q$,
we have $AS(\P \star_1 \Q) \subseteq AS(\P \star_2 \Q)$.
\label{prop:ordering-GLP}
\end{proposition}

This result paves the way for the choice of a specific GLP revision operator
depending on the desired ``amount of information'' provided by the revised GLP
in terms of number of its answer sets.
Precisely, any GLP revision operator $\star^{\circ, f}$ can be specified from an
answer set point of view by the following roadmap.
Since in the case where $\P + \Q$ is consistent, we always have $\P \star^{\circ, f} \Q = \P + \Q$,
the intuition underlying this procedure only applies when the programs
considered for the revision have no common SE model.
First, one
chooses a KM revision operator $\circ$ whose role is to filter the undesired answer
sets of the resulting revised program:
only the models $Y$ of the formula resulting from the revision of
$\P$ by $\Q$ in the propositional sense should be selected as
``potential answer set candidates''.
Then, the function $f$ plays a role in filtering those preselected candidates $Y$, so that $f$ can be defined
according to the following intuition: the more interpretations $X \subsetneq Y$ are included in $f(Y)$, the less likely the
interpretation $Y$ will actually be an answer set of the resulting revised program.
More precisely, the presence of a given interpretation $X \subsetneq Y$ in
$f(Y)$ is enough to discard $Y$ as being an answer set of the resulting revised program when $(X, Y)$ is an SE model of $\Q$.

This brings in light that, depending on the ``position'' of the GLP revision operator $\star^{\circ, f}$ in the lattice
$(\GLPops{\circ}, \lattice{\circ})$, when revising $\P$ by $\Q$ one may expect divergent results for $AS(\P \star^{\circ, f} \Q)$.
%
%
%
%
%
We illustrate this claim by considering
two specific classes of GLP revision operators that correspond respectively
to the suprema and infima of lattices $(\GLPops{\circ}, \lattice{\circ})$ for all
KM revision operators \nolinebreak $\circ$. The first ``extreme'' operators
are defined as follows:

\begin{definition}[Skeptical GLP revision operators]
The \emph{skeptical GLP revision operators}, denoted $\star^\circ_S$
are the propositional-based GLP revision operators $\star^{\circ, f}$
where $f$ is defined for every interpretation $Y$ by $f(Y) = 2^Y$.
\label{def:cautious operator}
\end{definition}

Note that skeptical GLP revision operators include the drastic GLP revision operator $\star_D$ (cf. Definition \ref{def:drastic}),
i.e., $\star_D = \star^{\circ_D}_S$ where $\circ_D$ is the (propositional) drastic revision operator.
For each propositional revision operator $\circ$, we clearly have
$\star^\circ_S = inf(\GLPops{\circ}, \lattice{\circ})$. We provide
now an axiomatic characterization of the skeptical GLP revision operators
in order to get a clearer view of their general behavior:

\begin{proposition}
The skeptical GLP revision operators are the only GLP revision operators $\star$ such that
for all GLPs $\P, \Q$,
whenever $\P + \Q$ is inconsistent, we have $AS(\P \star \Q) \subseteq AS(\Q)$.
\label{prop:sceptical}
\end{proposition}

Remark that the drastic GLP revision operator (cf. Definition \ref{def:drastic}), i.e.,
the skeptical GLP revision operator based on the propositional drastic revision operator $\star^{\circ_D}_S$,
is a specific case from the result given in Proposition \ref{prop:sceptical} where $AS(\P \star^{\circ_D}_S \Q) = AS(\Q)$
whenever $\P + \Q$ is inconsistent.

We now introduce another class of GLP revision operators which correspond to the other ``extreme cases''
with respect to lattices $(\GLPops{\circ}, \lattice{\circ})$:

\begin{definition}[Brave GLP revision operators]
The \emph{brave GLP revision operators}, denoted $\star^\circ_B$
are the propositional-based GLP revision operators $\star^{\circ, f}$
where $f$ is defined for every interpretation $Y$ by $f(Y) = \{Y\}$.
\label{def:brave operator}
\end{definition}


We get now that for each propositional revision operator $\circ$,
$\star^\circ_B = sup(\GLPops{\circ},$\linebreak$\lattice{\circ})$. The brave operators
are axiomatically characterized as follows:

\begin{proposition}
The brave GLP revision operators are the only GLP revision operators $\star^{\circ, f}$ such that
for all GLPs $\P, \Q$, whenever $\P + \Q$ is inconsistent, we have
$AS(\P \star^{\circ, f} \Q) = mod(\formGLPb{\P} \circ \formGLPb{\Q})$.
\label{prop:brave}
\end{proposition}

Let us remark as a specific case that the brave GLP revision operator based on the propositional drastic revision operator,
i.e., the operator $\star^{\circ_D}_B$, satisfies $AS(\P \star^{\circ_D}_B \Q) = mod(\Q)$ whenever $\P + \Q$ is inconsistent.

The following representative example illustrates how much the behavior
of skeptical and brave GLP revision operators diverge:

\begin{example}
Consider $\circ_D$, i.e., the propositional drastic revision operator.
Let $\P =
\left\lbrace\begin{array}{l}
p \leftarrow \top,\\
q \leftarrow \top,\\
\bot \leftarrow r
\end{array}\right\rbrace$
and $\Q = \{\bot \leftarrow p, q, \sim{r}\}$.
We have $AS(\P) = \{p, q\}$, $AS(\Q) = \{\emptyset\}$, and
$$\left\lbrace\begin{array}{l}
AS(\P \star^{\circ_D}_S \Q) = \{\emptyset\},\\
AS(\P \star^{\circ_D}_B \Q) = \{\emptyset, p, q, r, pr, qr, pqr\}.
\end{array}\right.$$
\end{example}

Though they are rational LP revision operators w.r.t. the postulates (RA1 - RA6),
skeptical and brave operators have a rather trivial, thus undesirable behavior.
Consider first the case of skeptical operators and assume that
the proposition $p$ is believed to be false, then learned to be true. That is,
$\{\bot \leftarrow p\} \subseteq \P$ and $\Q = \{p \leftarrow \top\}$.  Then one obtains that
$AS(\P \star^{\circ}_S \Q) \subseteq AS(\Q)$, that is, $AS(\P \star^{\circ}_S \Q) \subseteq \{p\}$, i.e.,
for any such program $\P$, on learning that $p$ is true the revision states that \emph{only} $p$ may be true,
which holds independently from the choice of the KM revision operator $\circ$.
On the other hand, brave operators only focus on classical models of logic programs $\P, \Q$ to compute
$\P \star^{\circ}_B \Q$ (whenever $\P + \Q$ is inconsistent), thus they do not take into consideration
the inherent, non-monotonic behavior of logic programs. As a consequence, programs $\P \star^{\circ}_B \Q$
will often admit many answer sets that are actually irrelevant to the input programs $\P$ and $\Q$.

Stated otherwise, skeptical and brave GLP revision operators are dual sides of a ``drastic'' behavior for the revision.
These operators are representative examples that provide some ``bounds'' of the complete picture
of GLP revision operators $\GLPops{\circ}$, for each KM revision operator $\circ$.
Discarding such drastic behaviors may call for additional postulates in order to capture
more parsimonious revision procedures in logic programming, as for instance the
cardinality-based revision operator (cf. Definition \ref{def:cardinal-based})
which is neither brave nor skeptical.
Then it seems necessary to refine the existing properties
that every rational revision operator
should satisfy so that the answer sets of the revised program $\P \star^{\circ, f} \Q$ fall ``between''
these two extremes (i.e., between $AS(\Q)$ and $mod(\P \circ \Q)$ in the sense of set inclusion).\\

Another benefit from our characterization result is that one can easily derive computational results by exploiting existing ones from propositional revision.
We assume that the reader is familiar with the basic concepts of computational complexity, in particular
with the classes $\mathbf{P}$, $\mathbf{NP}$ and co$\mathbf{NP}$
(see \cite{1994-papadimitriou} for more details).
Higher complexity classes are defined using oracles.
In particular $\mathsf{P}^{\mathsf{C}}$ corresponds to the class of decision problems that are solved
in polynomial time by deterministic Turing machines using an oracle for $\mathsf{C}$ in polynomial time.
For instance, $\Theta^p_2 = \mathbf{P}^{\mathbf{NP}[{\cal O}(log \ n)]}$
is the class of problems that can be solved in polynomial time by a deterministic Turing machine using a number of calls
to an $\mathbf{NP}$ oracle bounded by a logarithmic function of the size of the input representation of the problem.


We focus here on the the \emph{model-checking problem} \cite{DBLP:journals/jcss/LiberatoreS01}
for LP revision operators.
In the propositional case, the model-checking problem consists in
deciding whether a (propositional) interpretation is supported by a revised formula:
\begin{problem}[$\mathsf{MC}(\circ)$]
\begin{itemize}
\item {\bf Input:} 
\emph{A propositional revision operator $\circ$, two formulae $\phi, \psi$ and an interpretation $I$,}
\item {\bf Question:} 
\emph{Does $I \models \phi \circ \psi$ hold?}
\end{itemize}
\label{pb:mc-prop}
\end{problem}
The model-checking problem for the drastic revision operator (cf. Definition \ref{def:drasticr}) is co$\mathsf{NP}$-complete, while it is
$\Theta^p_2$-complete for the Dalal revision operator (cf. Definition \ref{def:dalal}):
\begin{proposition}
$\mathsf{MC}(\circ_D)$ is co$\mathsf{NP}$-complete.
\label{prop:drastic coNP}
\end{proposition}
\begin{theorem}[Liberatore and Schaerf 2001]
$\mathsf{MC}(\circ_{Dal})$ is $\mathsf{\Theta^p_2}$-complete.
\label{th:dalal theta2p}
\end{theorem}
Similarly one can consider the
model-checking problem for LP revision operators which consists in deciding whether an SE interpretation is an
SE model of a revised program:
\begin{problem}[$\mathsf{MC}_{\cal SE}(\star)$]
\begin{itemize}
\item {\bf Input:} 
\emph{An LP revision operator $\star$, two GLPs $\P, \Q$ and an SE interpretation $(X, Y)$,}
\item {\bf Question:} 
\emph{Does $(X, Y)$ belong to $\SE(\P \star \Q)$?}
\end{itemize}
\label{pb:mc-prop}
\end{problem}
Remark that given an SE interpretation $(X, Y)$ and a logic program $\P$, checking whether $(X, Y)$ is an SE model of
$\P$ is in $\mathbf{P}$: computing the program $\P^Y$, i.e., the reduct of $\P$ relative to $Y$, is performed in polynomial time;
then it is enough to check whether $Y \models \P$ and $X \models \P^{Y}$ which is performed in polynomial time.
Interestingly, when $f$ is computed in polynomial time the model-checking problem for propositional-based LP revision operators $\star^{\circ, f}$ is not harder
than the counterpart problem for the propositional revision operator $\circ$. Obviously enough, this applies for both skeptical and brave GLP revision operators, so
Proposition \ref{prop:drastic coNP} and Theorem \nolinebreak\ref{th:dalal theta2p} provide us with the following complexity results:

\begin{corollary}
\begin{itemize}
\item $\mathsf{MC}_{\cal SE}(\star^{\circ_{D}}_S)$ and $\mathsf{MC}_{\cal SE}(\star^{\circ_{D}}_B)$ are co$\mathsf{NP}$-complete;
\item $\mathsf{MC}_{\cal SE}(\star^{\circ_{Dal}}_S)$ and $\mathsf{MC}_{\cal SE}(\star^{\circ_{Dal}}_B)$ are $\mathsf{\Theta^p_2}$-complete.
\end{itemize}
\end{corollary}


\section{The case of disjunctive and normal logic programs}
\label{sec:disjunctive-normal}

In this section we take a look at more restrictive forms of programs, i.e., the disjunctive logic programs
and the normal logic programs.
A \emph{disjunctive logic program} (DLP) is a GLP where rules are of the form
$$\begin{array}{l}
a_1; \dots; a_k \leftarrow b_1, \dots, b_l, \sim c_1, \dots, \sim c_m,
\end{array}\trichea$$
where $k, l, m \geq 0$.
A \emph{normal logic program} (NLP) is a DLP where $k = 1$.

Let us recall that every GLP has a \emph{well-defined} set $S$ of SE models, which requires that $(Y, Y) \in S$ for every $(X, Y) \in S$,  and that
conversely, for every well-defined set $S$ of SE interpretations one can build a GLP $P$ such that $\SE(\P) = S$.
Since NLPs and DLPs are syntactically more restrictive than GLPs, these programs are characterized by sets of SE models satisfying
stronger conditions. A set of SE interpretations $S$ is said to be:
\begin{itemize}
\item \emph{complete} if it is well-defined and for all interpretations $X, Y, Z$, if $Y \subseteq Z$ and $(X, Y), (Z, Z) \in S$ then also $(X, Z) \in S$;
\item \emph{closed under here-intersection} if it is complete and for all interpretations $X, Y, Z$, if $(X, Z), (Y, Z) \in S$ then also $(X \cap Y, Z) \in S$.
\end{itemize}
Each DLP (respectively, NLP) has a complete (respectively, closed under here-intersection) set of SE models. Conversely, if a set of SE interpretations
$S$ is complete (respectively, closed under here-intersection) then one can build a DLP (respectively, NLP) $\P$ such that $\SE(\P) = S$
\cite{Eiter05onsolution,Cabalar}.
For instance, one can easily check that:
\begin{itemize}
\item the logic program $\P =
\left\lbrace\begin{array}{l}
p \leftarrow \sim q,\\
\bot \leftarrow p, q
\end{array}\right\rbrace$
from Example \ref{ex:detailed GLP P} is a NLP and $\SE(\P)$ is well-defined, complete and closed under here-intersection;
\item the logic program $\P_2 =
\left\lbrace\begin{array}{l}
p \leftarrow \sim{q},\\
p; q \leftarrow \top
\end{array}\right\rbrace$ from Example \ref{ex:difference-asp-se} is a DLP and $\SE(\P_2)$ is well-defined and complete,
but not closed under here-intersection;
\item the logic program $\Q_2 =
\left\lbrace\begin{array}{ll}
\bot \leftarrow p, \sim q, & \bot \leftarrow q, \sim p,\\
p; \sim{p} \leftarrow \top, & q; \sim{q} \leftarrow \top.
\end{array}\right\rbrace$ from Example \ref{ex:first-GLP-revision} is a GLP and $\SE(\Q)$ is well-defined but not complete.
\end{itemize}

When revising a logic program by another one, one expects the resulting revised program to be expressed in the same language
as the input programs.

\begin{definition}[DLP/NLP revision operator]
A \emph{DLP revision operator} (respectively, a \emph{NLP revision operator}) $\revision$ is an LP revision operator
associating two DLPs (respectively, two NLPs) $\P, \Q$ with a new DLP (respectively, a new NLP) $\P \revision \Q$,
and which satisfies postulates (RA1 - RA6).
\label{def:DLP-NLP-revision-operator}
\end{definition}

We first remark that both sets of DLP revision operators and NLP revision operators are not empty. Indeed, one can observe that
the intersection of two complete sets of SE interpretations is also complete, thus the expansion of two DLPs
leads to a DLP. This also applies for NLPs. As a direct consequence,
the drastic LP revision operator (cf. Definition \ref{def:drastic})
is both a DLP revision operator and a NLP revision operator. In fact, we have the more
general result:

\begin{proposition}
The skeptical GLP revision operators are both DLP revision operators and NLP revision operators.
\label{prop:skeptical-DLP-NLP}
\end{proposition}

However, the above result does not apply for all GLP revision operators. That is to say, there exist
some GLP revision operators which associate two NLPs with a GLP which is not a DLP. Hence,
our sound and complete construction of GLP revision operators does not hold anymore for DLP and NLP
revision operators.
For instance, brave GLP revision operators are neither DLP revision operators nor NLP revision operators,
as shown in the following example:

\begin{example}
Let $\P =
\left\lbrace\begin{array}{l}
\bot \leftarrow \sim{p}, \sim{q},\\
\bot \leftarrow q, \sim{p},\\
\bot \leftarrow p, q
\end{array}\right\rbrace$
and $\Q = \{q \leftarrow \top\}$ be two NLPs. We have that $\SE(\P) = \{(\emptyset, p), (p, p)\}$ and
$\SE(\Q) = \{(q, q), (q, pq), (pq, pq)\}$.
Consider the brave GLP revision operator $\star^{\circ_D}_B$ based on the
propositional drastic revision operator. Then one can verify that
$\SE(\P \star^{\circ_D}_B \Q) = \{(q, q), (pq, pq)\}$ is not a complete set of SE interpretations,
thus $\P \star^{\circ_D}_B \Q$ cannot be represented as a DLP.
\end{example}

As a consequence, our characterization result from Proposition \ref{prop:characterization-revision-GLP}
does not hold anymore for DLP/NLP revision operators.
Nevertheless, we provide below a representation of both DLP and NLP revision operators
in terms of two structures where the first one is an LP faithful assignment adapted to DLPs/NLPs and the second one
is a well-defined assignment ``strengthened'' by some further conditions.

\begin{definition}[DLP/NLP faithful assignment]
A \emph{DLP faithful assignment} (respectively, a \emph{NLP faithful assignment}) is a mapping which associates every DLP
(respectively, every NLP) with
a total preorder over interpretations such that
conditions (1 - 3) of an LP faithful assignment are satisfied.
\label{def:DLPNLPfaithful}
\end{definition}

\begin{definition}[Complete assignment]
Let $\Phi$ be a DLP faithful assignment which associates every DLP $\P$ with a total
preorder $\leq_\P$.
A \emph{$\Phi$-based complete assignment}
is a mapping which associates with every DLP $\P$ and
every interpretation $Y$ a set of interpretations denoted by $\P_\Phi(Y)$, such that conditions (a - e)
of a well-defined assignment are satisfied as well as the following further condition, for all interpretations $X, Y, Z$:
\begin{description}
\item[(f)] If $X \in \P_\Phi(Y)$, $Y \simeq_\P Z$ and $Y \subseteq Z$ then $X \in \P_\Phi(Z)$.
\end{description}
A pair $(\Phi, \Psi_\Phi)$, where $\Phi$ is a DLP faithful assignment and $\Psi_\Phi$ is a $\Phi$-based complete assignment,
is called a \emph{DLP parted assignment}.
\label{def:DLP-parted-assignment}
\end{definition}

\begin{definition}[Normal assignment]
Let $\Phi$ be a NLP faithful assignment.
A \emph{$\Phi$-based normal assignment}
is a mapping which associates with every NLP $\P$ and
every interpretation $Y$ a set of interpretations denoted by $\P_\Phi(Y)$, such that conditions (a - f)
of a complete assignment are satisfied as well as the following further condition, for all interpretations $X, Y, Z$:
\begin{description}
\item[(g)] If $X, Y \in \P_\Phi(Z)$ then $X \cap Y \in \P_\Phi(Z)$.
\end{description}
A pair $(\Phi, \Psi_\Phi)$, where $\Phi$ is a NLP faithful assignment and $\Psi_\Phi$ is a $\Phi$-based normal assignment,
is called a \emph{NLP parted assignment}.
\label{def:NLP-parted-assignment}
\end{definition}

We are ready to provide our characterization results for DLP revision operators and NLP revision operators:

\begin{proposition}
An LP operator $\star$ is a DLP (resp. NLP) revision operator
if and only if there exists a DLP (resp. NLP) parted assignment $(\Phi, \Psi_\Phi)$, where
$\Phi$ associates with every DLP (resp. NLP) $\P$ a total preorder $\leq_\P$,
$\Psi_\Phi$ is a $\Phi$-based complete (resp. normal) assignment
which associates with every DLP (resp. NLP) $\P$ and every interpretation $Y$ a set of interpretations $\P_\Phi(Y)$, and
such that for all DLPs
(resp. NLPs)
$\P, \Q$,
$$\SE(\P \star \Q) = \{(X, Y) \mid (X, Y) \in \SE(\Q),
Y \in \min(\modeles{\Q}, \leq_\P),
X \in \P_\Phi(Y)\}.$$
\label{prop:last-result}
\end{proposition}

As to the case of our characterization of GLP revision operators,
Proposition \ref{prop:last-result} provides us with sound and complete constructions of DLP and NLP revision operators
in terms of total preorders over propositional interpretations and some further conditions specific to SE interpretations.
Furthermore, because both constructions are similar to the one of GLP revision operators, without stating it formally one can
straightforwardly establish a one-to-one correspondence between DLP/NLP revision operators and propositional-based
LP revision operators (cf. Definition \ref{def:propbasedGLPoperator}) satisfying some further conditions on the function $f$
very similar to conditions (f) and (g).
Indeed, one can see from Definition \ref{def:DLP-parted-assignment} and \ref{def:NLP-parted-assignment} that
the two structures involved in DLP/NLP parted assignments are not independent anymore,
since by condition (f) the $\Phi$-based complete and normal assignments
should both be aligned with the corresponding faithful assignment. As a consequence, these structures are more complex than those of GLP parted
assignments and similar embeddings of DLP/NLP revision operators into Boolean lattices are no more applicable.
A deeper investigation of the type of ruling structures for $\Phi$-based complete and normal assignments is out of the scope of this paper, but
constitutes an interesting direction to explore in a future work.

\section{Conclusion}
\label{sec:conclusion}

In this paper, we pursued some previous work on revision of logic programs, where
the adopted approach is based on a monotonic characterization of logic programs using
SE interpretations. We gave a particular attention to the revision of generalized logic programs (GLPs)
and characterized the class of rational GLP revision operators
in terms of total orderings among classical interpretations with some further conditions specific
to SE interpretations. The constructive characterization we provided
facilitates the comprehension of the semantic properties of GLP revision operators by drawing a clear, complete picture
of them.
Interestingly, we showed that a GLP revision operator can be viewed as an extension of
a rational propositional revision operator: each propositional revision operator corresponds to a specific subclass of GLP
revision operators, and a GLP revision operator from a particular subclass can be specified independentely of the propositional revision operator
under consideration.
Moreover, we showed that each one of these subclasses can be embedded into a Boolean lattice
whose infimum and supremum, the so-called \emph{skeptical} and \emph{brave} GLP revision operators, have some relatively drastic behavior.
In addition, we adjusted our representation structures and provided sound and complete constructions for two more
specific classes of logic programs, i.e., the disjunctive and normal logic programs.

Our results make easier the improvement of the current AGM framework in the context of logic programming.
Indeed, though the subclasses of skeptical and brave revision operators are fully satisfactory w.r.t. the AGM revision principles,
their behavior is shown to be rather trivial. This may call for additional postulates which would aim to capture more parsimonious,
``balanced'' classes of revision operators.

As to the case of \emph{update} of logic programs
Slota and Leite \citeyear{DBLP:journals/corr/SlotaL13} argued that semantic rule updates based on SE models seem to be inappropriate. Indeed they showed
that in presence of the irrelevance-of-syntax postulate (whose counterpart in the context of revision is (RA4)), semantic rule
update operators based on SE models violate some reasonable properties for rule updates, i.e., \emph{dynamic support} and \emph{fact update}
(see \cite{DBLP:journals/corr/SlotaL13} for more details). The property of dynamic support can be expressed unformally as follows:
an rule update operator $\oplus$ satisfies dynamic support if every atom true in an answer set from any updated program $\P \oplus \Q$
should be supported by a rule in $\P \cup \Q$, i.e., it should have some ``justification'' in either the original program or the new one.
The property of fact update requires some notion of atom inertia when updating 
a consistent set of facts (i.e., a set of rules of the type $p \leftarrow \top$ where $p$ is an atom)
by a consistent set of facts.
Both of these properties require rule update operators to have a reasonable ``syntactic'' behavior, away from the purely semantic approach
represented by the adapted AGM postulates. In \cite{DBLP:conf/kr/SlotaL12} the same authors successfully reconciliate
semantic-based and syntax-based approaches to updating logic programs: they considered different characterizations of logic programs
in terms of RE models (standing for \emph{robust equivalence models}) that proved to be a more suitable semantic fundation for rule updates than SE models.
A straightforward direction of research is to investigate whether these richer characterizations of logic programs suit to revision operators.

Additionally, we will investigate the case of logic program \emph{merging} operators (merging can be viewed
as a multi-source generalization of belief revision, see for instance \cite{KP02}). Indeed it is not even known
whether there exists a fully rational merging operator, i.e., that satisfies the whole set of postulates
proposed by Delgrande \textit{et al.\/} \citeyear{DBLP:conf/iclp/DelgrandeSTW09,Delgrande/ACM} for logic program merging operators based on SE models.


\section*{Appendix: Proofs of Propositions}




%

\noindent {\it Proposition \ref{rem:faithful}}~\\
There is a one-to-one correspondence between the KM revision operators
and the set of all faithful assignments.

\vspace{2mm}

\begin{proof}
Let $\circ_1, \circ_2$ be two KM revision operators. From Theorem \ref{proposition : AGM revision operator caracterization}
one can build two faithful assignments associating respectively with every formula $\phi$ the total preorders $\leq^1_\phi$ (for the
first faithful assignment) and $\leq^2_\phi$ (for the second one),
such that for all formulae $\phi, \psi$, $\modeles{\phi \circ_1 \psi} = \min(\modeles{\psi}, \leq^1_\phi)$
and $\modeles{\phi \circ_2 \psi} = \min(\modeles{\psi},$ \linebreak $\leq^2_\phi)$.
Assume now that $\circ_1 \neq \circ_2$.
This means that there exist two propositional formulae $\phi, \psi$ such that
$\phi \circ_1 \psi \not\equiv \phi \circ_2 \psi$, so $\modeles{\phi \circ_1 \psi} \neq \modeles{\phi \circ_2 \psi}$,
thus $\min(\modeles{\psi}, \leq^1_\phi) \neq \min(\modeles{\psi}, \leq^2_\phi)$.
Hence, $\leq^1_\phi \neq \leq^2_\phi$, so the two faithful assignments associated respectively with $\circ_1$ and $\circ_2$
are different.
Conversely, assume that the two faithful assignments associated respectively with $\circ_1$ and $\circ_2$
are different. Then, there exists a formula $\phi$ such that $\leq^1_\phi \neq \leq^2_\phi$.
This means that there exists two interpretations $I, J$ such that $I \leq^1_\phi J$ and $J <^2_\phi I$.
Let $\psi$ be any formula such that $\modeles{\psi} = \{I, J\}$. We have $I \in \min(\modeles{\psi}, \leq^1_\phi)$
and $I \notin \min(\modeles{\psi}, \leq^2_\phi)$. Hence, $\modeles{\phi \circ_1 \psi} \neq \modeles{\phi \circ_2 \psi}$,
or equivalently, $\phi \circ_1 \psi \not\equiv \phi \circ_2 \psi$. This means that $\circ_1 \neq \circ_2$.
\end{proof}

%
%

\noindent {\it Proposition \ref{prop:drastic revision operator}}~\\
$\revision_D$ is a GLP revision operator.

\vspace{2mm}

\begin{proof}
Let $\P, \Q$ be two logic programs.
The fact that $\P \revision_D \Q$ returns a GLP
when $\P, \Q$ are both GLPs
is obvious from the definition.
Postulates (RA1 - RA4) are directly satisfied from the definition.
(RA5 - RA6) Let $\P, \Q, \R$ be three GLPs.
If $(\P \revision_D \Q) + \R$ is not consistent then (RA5) is trivially satisfied,
so assume that $(\P \revision_D \Q) + \R$ is consistent. We have to show that
$(\P \revision_D \Q) + \R \equiv_s \P \revision_D (\Q + \R)$.
We fall now into two cases. Assume first that $\P + \Q$ is consistent.
By definition, $(\P \revision_D \Q) + \R = \P + \Q + \R$. Yet since $(\P \revision_D \Q) + \R$ is consistent,
so is $\P + \Q + \R$, thus we get by definition $\P \revision_D (\Q + \R) = \P + \Q + \R$.
Therefore, $(\P \revision_D \Q) + \R \equiv_s \P \revision_D (\Q + \R)$.
Now, assume that $\P + \Q$ is not consistent. By definition, $(\P \revision_D \Q) + \R = \Q + \R$.
Since $\P + \Q$ is not consistent, we also have $\P + \Q + \R$ not consistent.
So by definition $\P \revision_D (\Q + \R) = \Q + \R$. Hence,
$(\P \revision_D \Q) + \R \equiv_s \P \revision_D (\Q + \R)$.
\end{proof}


\noindent {\it Proposition \ref{prop:characterization-revision-GLP}}~\\
An LP operator $\star$ is a GLP revision operator if and only if there exists a pair $(\Phi, \Psi)$, where
$\Phi$ is an LP faithful assignment associating with every GLP $\P$ a total preorder $\leq_\P$,
$\Psi$ is a well-defined assignment associating with every GLP $\P$ and every interpretation
$Y$ a set of interpretations $\P(Y)$, and such that for all GLPs $\P, \Q$,
$$\SE(\P \star \Q) = \{(X, Y) \mid (X, Y) \in \SE(\Q),
Y \in \min(\modeles{\Q}, \leq_\P),
X \in \P(Y)\}.$$

\vspace{2mm}

\begin{proof}
\textit{(Only if part)} In this proof, for every well-defined set of SE interpretations $S$, $\formwithoutbraces{S}$
denotes any GLP $\P$ such that $\SE(\P) = S$.
To alleviate notations, when $S$ is of the form $\{(Y, Y) \mid Y \in E\}$ for some set of interpretations $E$,
we write $\formwithoutbraces{E}$ instead of $\formwithoutbraces{S}$. For instance, $\form{(Y, Y), (Y', Y'), (Y^{(2)}, Y^{(2)})}$ will simply
be denoted by $\formYYpYpp$.
The proof exploits on several occasions the following remarks:

\begin{remark}
If $\star$ is an LP revision operator satisfying the postulates (RA5) and (RA6), then for all GLPs $\P, \Q, \R$
such that $(\P \star \Q) + \R$ is consistent, we have $(\P \star \Q) + \R \equiv_s \P \star (\Q + \R)$.
\label{rem:rem3}
\end{remark}

\begin{remark}
For all sets of interpretations
$E, F$, $\formwithoutbraces{E} + \formwithoutbraces{F} \equiv_s \formwithoutbraces{E \cap F}$.
\label{rem:rem2}
\end{remark}

\begin{remark}
Let $\star$ be an LP revision operator satisfying the postulates (RA1) and (RA3). Then for any GLP $\P$ and any non-empty set of interpretations $E$,
$mod(\P \star \formwithoutbraces{E}) \neq \emptyset$ and
$mod(\P \star \formwithoutbraces{E}) \subseteq E$.
\label{lemmaRA1RA3GLP}
\end{remark}

Let $\revision$ be a GLP revision operator.
For every GLP $\P$,
define the relation $\leq_\P$ over interpretations such that $\forall Y, Y' \in \allinterpretations$, $Y \leq_\P Y'$
iff $Y \models \P \revision \formYYp$. Moreover, for every GLP $\P$,
$\forall Y \in \allinterpretations$, let $\P(Y) = \{X \subseteq Y \mid \XY \in \SE(\P \revision \formXYY)\}$.
Let $\P$ be any GLP.
We first show that $\leq_\P$ is a total preorder.
Let
$Y, Y', Y^{(2)} \in \allinterpretations$.\\
\noindent (Totality of $\leq_\P$): By Remark \ref{lemmaRA1RA3GLP},
$Y \models \P \revision \formYYp$ or $Y' \models \P \revision \formYYp$.
Hence, $Y \leq_\P Y'$ or $Y' \leq_\P Y$.\\
\noindent (Reflexivity of $\leq_\P$): By Remark \ref{lemmaRA1RA3GLP}, $Y \models \P \revision \formY$, so $Y \leq_\P Y$.\\
\noindent (Transitivity of $\leq_\P$): Assume towards a contradiction that $Y \leq_\P Y'$, $Y' \leq_\P \Yppsingle$ and $Y \not\leq_\P \Yppsingle$.
We consider two cases:\\
\noindent Case 1: $(\P \revision \formYYpYpp) + \formYYpp$ is consistent. Then we have
$$\begin{array}{l}
(\P \revision \formYYpYpp) + \formYYpp\\
\begin{array}{ll}
\ \ \equiv_s \P \revision (\formYYpYpp + \formYYpp) & \mbox{(by Remark \ref{rem:rem3})}\\
\ \ \equiv_s \P \revision \formYYpp & \mbox{(by Remark \ref{rem:rem2})}.
\end{array}
\end{array}$$
Since $Y \not\leq_\P \Yppsingle$, by definition of $\leq_\P$ we get that $Y \not\models \P \revision \formYYpp$,
hence $Y \not\models \P \revision \formYYpYpp$.
By Remark \ref{lemmaRA1RA3GLP}, there are two remaining cases:
\begin{itemize}
\item[(i)] $Y' \models \P \revision \formYYpYpp$. In this case, 
$(\P \revision \formYYpYpp) + \formYYp$ is consistent, so
$$\begin{array}{l}
(\P \revision \formYYpYpp) + \formYYp\\
\begin{array}{ll}
\ \ \equiv_s \P \revision (\formYYpYpp + \formYYp) & \mbox{(by Remark \ref{rem:rem3})}\\
\ \ \equiv_s \P \revision \formYYp & \mbox{(by Remark \ref{rem:rem2})}.
\end{array}
\end{array}$$
Since $Y \leq_\P Y'$, by definition of $\leq_\P$ we get that $Y \models \P \revision \formYYp$,
hence $Y \models \P \revision \formYYpYpp$. which contradicts the previous conclusion that $Y \not\models \P \revision \formYYpYpp$.
\item[(ii)] $Y' \not\models \P \revision \formYYpYpp$. Since we also have that $Y \not\models \P \revision \formYYpYpp$,
by Remark \ref{lemmaRA1RA3GLP}
we must have that $\Yppsingle \models \P \revision \formYYpYpp$
In this case, $(\P \revision \formYYpYpp) + \formYpYpp$ is consistent, so
$$\begin{array}{l}
(\P \revision \formYYpYpp) + \formYpYpp\\
\begin{array}{ll}
\ \ \equiv_s \P \revision (\formYYpYpp + \formYpYpp) & \mbox{(by Remark \ref{rem:rem3})}\\
\ \ \equiv_s \P \revision \formYpYpp & \mbox{(by Remark \ref{rem:rem2})}.
\end{array}
\end{array}$$
Since $Y' \leq_\P \Yppsingle$, by definition of $\leq_\P$ we get that $Y' \models \P \revision \formYpYpp$,
hence $Y' \models \P \revision \formYYpYpp$, which is a contradiction.
\end{itemize}

\noindent Case 2: $(\P \revision \formYYpYpp) + \formYYpp$ is not consistent.
Then by Remark \nolinebreak\ref{lemmaRA1RA3GLP}, $Y' \models \P \revision \formYYpYpp$.
Then $(\P \revision \formYYpYpp) + \formYYp$ is consistent,
and by using Remark \ref{rem:rem3} and \ref{rem:rem2} and following similar reasonings as in (i),
we get that $Y' \models \P \revision \formYYp$ and $Y \not\models \P \revision \formYYp$.
By definition of $\leq_\P$ this contradicts $Y \leq_\P Y'$
and concludes the proof that 
$\leq_\P$ is a total preorder.

Now, let $\Q$ be any GLP.
We have to show that $\SE(\P \revision \Q) =
\{(X, Y) \mid (X, Y) \in \SE(\Q),
Y \in \min(\modeles{\Q}, \leq_\P),
X \in \P(Y)\}$. Let us denote by
$\S$ the latter set and first show the first inclusion $\SE(\P \revision \Q) \subseteq_s \S$.
Let $\XY \in \SE(\P \revision \Q)$ and let us show that (i) $\XY \in \SE(\Q)$,
(ii) $\forall Y' \models \Q, Y \leq_\P Y'$ and that (iii) $X \in \P(Y)$.
(i) is direct from (RA1). For (ii), let $Y' \models \Q$. Since $\revision$ returns a GLP, $\SE(\P \revision \Q)$ is well-defined.
That is, since $\XY \in \SE(\P \revision \Q)$, we have $Y \models \P \revision \Q$. Therefore,
$(\P \revision \Q) + \formYYp$ is consistent. So by Remark \ref{rem:rem3} and \ref{rem:rem2}, $Y \models \P \revision \formYYp$.
Hence, $Y \leq_\P Y'$.
For (iii), since $\XY \in \SE(\P \revision \Q)$, $(\P \revision \Q) + \formXYY$ is consistent,
so we have $\XY \in \SE(\P \revision \formXYY)$ by Remark \ref{rem:rem3} and \ref{rem:rem2}; hence, $X \in \P(Y)$.
Let us now show the other inclusion $\S \subseteq_s \SE(\P \revision \Q)$. Assume $\XY \in \S$.
Then $\forall Y' \models \Q$, $Y \leq_\P Y'$ and $X \in \P(Y)$. First, from the definition of $\P(Y)$
we have $Y \in \P(Y)$, so also $(Y, Y) \in \S$. Since $\S \neq \emptyset$, $\Q$ is consistent,
thus by Remark \ref{lemmaRA1RA3GLP} there exists $Y_* \models \Q$, $Y_* \models \P \revision \Q$.
Let $\Rsharp = \form{\XY, \Y, \Yetoile}$. Note that $\Rsharp \subseteq_s \Q$ and that
$(\P \revision \Q) + \Rsharp$ is consistent since $Y_*$ is a model of both $\P \revision \Q$ and $\Rsharp$.
Then by Remark \nolinebreak\ref{rem:rem3}
we get that
$(\P \revision \Q) + \Rsharp \equiv_s \P \revision (\Q + \Rsharp) \equiv_s \P \revision \Rsharp$.
Since we have to show that $(X, Y) \in \SE(\P \revision \Q)$, it comes down to show that
$\XY \in \SE(\P \revision \Rsharp)$. Assume towards a contradiction that
$\XY \notin \SE(\P \revision \Rsharp)$. By Remark \ref{lemmaRA1RA3GLP} and since
$Y_* \models \P \revision \Rsharp$, we have two cases:
(i) $Y \not\models \P \revision \Rsharp$. Since $(\P \revision \Rsharp) + \form{\Y, \Yetoile}$ is consistent,
by Remark \ref{rem:rem3} and \ref{rem:rem2} we get that
$Y \not\models \P \revision \form{\Y, \Yetoile}$. This contradicts $Y \leq_\P Y_*$.
(ii) $Y \models \P \revision \Rsharp$. Since $(\P \revision \Rsharp) + \formXYY$ is consistent,
by Remark \ref{rem:rem3} and \ref{rem:rem2} we get that $(X, Y) \notin \SE(\P \revision \formXYY)$.
This contradicts $X \in \P(Y)$.


It remains to verify that all conditions (1 - 3) of the faithful assignment and conditions (a - e) of the
well-defined assignment are satisfied:\\
\begin{itemize}
\item[(1)] Assume $Y \models \P$ and $Y' \models \P$. By (RA2), $\P \revision \formYYp \equiv_s \P + \formYYp$.
So $Y \models \P \revision \formYYp$ and $Y' \models \P \revision \formYYp$,
hence $Y \simeq_\P Y'$;
\item[(2)] Assume $Y \models \P$ and $Y' \not\models \P$. By (RA2), $\P \revision \formYYp \equiv_s \P + \formYYp$.
So $Y \models \P \revision \formYYp$ and $Y' \not\models \P \revision \formYYp$,
hence $Y <_\P Y'$;
\item[(3)] Obvious from (RA4);
\item[(a)] By definition of $\P(Y)$ and by (RA1) and (RA3), we must have $Y \models \P \revision \formXYY$, i.e., $Y \models \P(Y)$;
\item[(b)] If $X \in \P(Y)$ then $X \subseteq Y$ by definition of $\P(Y)$;
\item[(c)] Assume $\XY \in \SE(\P)$. Then $Y \models \P$.
By (RA2), $\P \revision \formXYY \equiv_s \P + \formXYY \equiv_s \formXYY$, so $\XY \in \SE(P \revision \formXYY)$.
Therefore, $X \in \P(Y)$.
\item[(d)] Assume $\XY \notin \SE(\P)$ and $Y \models \P$. By (RA2), $\P \revision \formXYY \equiv_s \P + \formXYY \equiv_s \formY$,
so $\XY \notin \formXYY$. Therefore, $X \notin \P(Y)$.
\item[(e)] Obvious from (RA4).\\
\end{itemize}

\noindent\textit{(If part)} We consider a faithful assignment that associates with every GLP $\P$ a total preorder $\leq_\P$
and a well-defined assignment that associates with every GLP $\P$ and every interpretation
$Y$ a set $\P(Y) \subseteq \allinterpretations$. For all GLPs $\P, \Q$, let $\S(\P, \Q)$ be the set of SE interpretations defined as
$\S(\P, \Q) =\{(X, Y) \mid (X, Y) \in \SE(\Q),
Y \in \min(\modeles{\Q}, \leq_\P),
X \in \P(Y)\}$. Let $\P, \Q$ be two GLPs and let us show that
$\S(\P, \Q)$ is well-defined. Let $(X, Y) \in \S(\P, \Q)$. By condition (a) of the well-defined assignment and since $X \subseteq Y$,
we have $Y \in \P(Y)$, so $(Y, Y) \in \S(\P, \Q)$. Hence, $\S(\P, \Q)$ is well-defined.
Then let us define an operator $\revision$ associating two
GLPs $\P, \Q$ with a new GLP $\P \revision \Q$ such that for all GLPs $\P, \Q$,
$\SE(\P \revision \Q) = \S(\P, \Q)$.

It remains to show that postulates (RA1 - RA6) are satisfied. Let $\P, \Q$ bet two GLPs.
\begin{itemize}
\item[(RA1)] By definition, $\SE(\P \revision \Q) \subseteq \SE(\Q)$.
\item[(RA2)] Assume that $\P + \Q$ is consistent. We have to show that $\SE(\P \revision \Q) = \SE(\P + \Q)$.
We first show the inclusion $\SE(\P \revision \Q) \subseteq \SE(\P + \Q)$. Let $(X, Y) \in \SE(\P \revision \Q)$.
Towards a contradicton, assume that $(X, Y) \notin \SE(\P + \Q)$. By definition of $\revision$ we have $(X, Y) \in \SE(\Q)$,
thus $(X, Y) \notin \SE(\P)$. We fall into two cases:\\
\noindent (i) $(Y, Y) \in \SE(\P)$. Then from condition (d), we have $X \notin \P(Y)$. This contradicts $(X, Y) \in \SE(\P \revision \Q)$;\\
\noindent (ii) $(Y, Y) \notin \SE(\P)$. Then from condition (2), $\forall Y' \models \P$, $Y' <_\P Y$. In particular,
$\forall Y' \models P + Q$, $Y' <_\P Y$. This contradicts $\XY \in \SE(\P \revision \Q)$.\\
We now show the other inclusion $\SE(\P + \Q) \subseteq \SE(\P \revision \Q)$. Let $(X, Y) \in \SE(\P + \Q)$.
So $(X, Y) \in \SE(\Q)$. From conditions (1) and (2), $\forall Y' \in \allinterpretations$, $Y <_\P Y'$.
Moreover from condition (c), since $\XY \in \SE(\P)$ we get that $X \in \P(Y)$. Therefore, $\XY \in \SE(\P \revision \Q)$.
\item[(RA3)] Suppose that $Q$ is consistent, i.e., $\SE(Q) \neq \emptyset$. As $\allinterpretations$ is a finite set of interpretations,
we have no infinite descending chain of inequalities w.r.t.~$\leq_\P$. Moreover, $\leq_\P$ is a total relation.
Hence, there is an interpretation $Y_* \models \Q$ such that $\forall Y' \models \Q$, $Y_* \leq_\P Y'$.
Lastly by condition (a), $Y_* \in \P_{Y_*}$. Hence, $Y_* \models \P \revision \Q$, i.e., $\P \revision \Q$ is consistent.
\item[(RA4)] Obvious by definition of $\revision$ and from conditions (3) and (e).
\item[(RA5)] Let $(X, Y) \in \SE((\P \revision \Q) + \R)$. So by definition of $\revision$,
$\forall Y' \models \Q$,
$Y \leq_\P Y'$ and $X \in \P(Y)$. In particular, $\forall Y' \models \Q + \R$, $Y \leq_\P Y'$ and $X \in \P(Y)$.
So $\XY \in \SE(\P \revision (\Q + \R))$.
\item[(RA6)] Assume that $(\P \revision \Q) + \R$ is consistent. Let $Y_{*} \models (\P \revision \Q) + \R$.
Let $\XY \in \SE(\P \revision (\Q + \R))$. Assume towards a contradiction that $\XY \notin \SE((\P \revision \Q) + \R)$.
Since $\XY \in \SE(\R)$, we have $\XY \notin \SE(\P \revision \Q)$. But $\XY \in \SE(\Q)$,
this means that $Y_{*} <_\P Y$ or $X \notin \P(Y)$. Yet $Y_* \models \Q + \R$,
so $(X, Y) \notin \SE(\P \revision (\Q + \R))$. This leads to a contradiction.
\end{itemize}
\end{proof}

%
%
%


\noindent {\it Proposition \ref{prop:prop-based equl GLP}}~\\
An LP revision operator is a GLP revision operator if and only if it is a propositional-based GLP revision operator.

\vspace{2mm}

\begin{proof}
\noindent \textit{(Only If part)} Let $\star$ be a GLP revision operator. We have to show that there exists
a KM revision operator $\circ$ and a mapping $f$ from $\allinterpretations$ to $2^\allinterpretations$
such that $\forall Y \in \allinterpretations$, $Y \in f(Y)$ and if $X \in f(Y)$ then $X \subseteq Y$, and such that
for all GLPs $\P, \Q$, $\SE(\P \star \Q) = \SE(\P \star^{\circ, f} \Q)$.
Yet from Proposition
\ref{prop:characterization-revision-GLP} there exists a GLP parted assignment $(\Phi, \Psi)$, where
$\Phi$ associates with every GLP $\P$ a total preorder $\leq_\P$
and $\Psi$ associates with every GLP $\P$ and every interpretation
$Y$ a set of interpretations $\P(Y)$, such that for all GLPs $\P, \Q$,
$\SE(\P \star \Q) = \{(X, Y) \mid (X, Y) \in \SE(\Q), Y \in \min(\modeles{\Q}, \leq_\P), X \in \P(Y)\}$.
Then, let $\circ$ be the KM revision operator associated with the faithful assignment (cf. Definition \ref{def:faithful})
that associates with every propositional
formula $\phi$ the total preorder $\leq_\phi = \leq_\P$, where $\P$ is any GLP such that $\phi \equiv \formGLPb{\P}$
(from Remark \ref{remark}, such an assignment is, indeed, faithful and unique). Then from Theorem
\ref{proposition : AGM revision operator caracterization}, for every $Y \in \allinterpretations$, $Y \in \min(\modeles{\Q}, \leq_\P)$
if and only if $Y \models \formGLPb{\P} \circ \formGLPb{\Q}$.
Then define $f$ as the mapping from $\allinterpretations$ to $2^\allinterpretations$ such that $\forall Y \in \allinterpretations$,
$f(Y) = \P(Y)$. From conditions (a) and (b) of the well-defined assignment (cf. Definition \ref{def:well-defined-assignment}),
$f$ is such that $\forall Y \in \allinterpretations$, $Y \in f(Y)$ and if $X \in f(Y)$ then $X \subseteq Y$. Now, given two GLPs $\P, \Q$,
if $\P + \Q$ is consistent, we directly get $\SE(\P \star \Q) = \SE(\P \star^{\circ, f} \Q)$ from Definition \ref{def:propbasedGLPoperator} and postulate (RA2).
So assume that $\P + \Q$ is inconsistent.
Given an SE interpretation
$(X, Y)$, we have $(X, Y) \in \SE(\P \star \Q)$ if and only if $(X, Y) \in \SE(\Q)$, $Y \in \min(\modeles{\Q}, \leq_\P)$ and $X \in \P(Y)$,
if and only if $(X, Y) \in \SE(\Q)$, $Y \models \formGLPb{\P} \circ \formGLPb{\Q}$ and $X \in f(Y)$, if and only if
$(X, Y) \in \SE(\P \star^{\circ, f} \Q)$. That is to say, $\SE(\P \star \Q) = \SE(\P \star^{\circ, f} \Q)$.\\

\noindent \textit{(If part)} Let $\star^{\circ, f}$ be a propositional-based GLP revision operator. We have to show that there exists
a GLP revision operator $\star$ such that $\SE(\P \star^{\circ, f} \Q) = \SE(\P \star \Q)$.
Since $\circ$ is a KM revision operator, from Theorem \ref{proposition : AGM revision operator caracterization} there is a faithful assignment
associating with every propositional formula $\phi$ a total preorder $\leq_{\phi}$. Then using Remark \ref{remark},
let $\Phi$ be the LP faithful assignment associating with every GLP $\P$ the total preorder $\leq_{\P} = \leq_{\phi}$, where $\phi$ is any propositional
formula such that $\formGLPb{\P} \equiv \phi$. From Theorem \ref{proposition : AGM revision operator caracterization}, 
for all GLPs $\P, \Q$ and for every $Y \in \allinterpretations$, $Y \models \formGLPb{\P} \circ \formGLPb{\Q}$ if and only if $Y \in \min(\modeles{\Q}, \leq_\P)$.
Now, let $\Psi$ be the mapping associating with every GLP $\P$ and every interpretation
$Y$ the set of interpretations $\P(Y) = \{X \in \allinterpretations \mid (X, Y) \in \SE(\P)\} \cup \{X \in f(Y) \mid Y \not\models \P\}$.
By definition, $\Psi$ satisfies conditions (a) - (e) of a well-defined assignment (cf. Definition \ref{def:well-defined-assignment}).
Then, let us consider the GLP revision operator $\star$ associated with the GLP parted assignment $(\Phi, \Psi)$.
We need to check that for all GLPs $\P, \Q$, $\SE(\P \star \Q) = \SE(\P \star^{\circ, f} \Q)$. Given two GLPs $\P, \Q$, if $\P + \Q$ is consistent,
we directly get $\SE(\P \star \Q) = \SE(\P \star^{\circ, f} \Q)$ from Definition \ref{def:propbasedGLPoperator} and postulate (RA2). So assume
that $\P + \Q$ is inconsistent. We first prove that $\SE(\P \star \Q) \subseteq \SE(\P \star^{\circ, f} \Q)$. Let $(X, Y) \in \SE(\P \star \Q)$.
We have $(X, Y) \in \SE(\Q)$, $Y \in \min(\modeles{\Q}, \leq_\P)$ and $X \in \P(Y)$.
Thus $(X, Y) \in \SE(\Q)$, $Y \models \formGLPb{\P} \circ \formGLPb{\Q}$ and $X \in \P(Y)$. We need to show that $X \in f(Y)$.
Yet since $\P + \Q$ is inconsistent, we have $(X, Y) \not\in \SE(\P)$; and since $(X, Y) \in \SE(\Q)$, we also have $(Y, Y) \in \SE(\Q)$,
so $(Y, Y) \not\in \SE(\P)$, thus $Y \not\models \P$. By definition of $\P(Y)$, this means that $X \in f(Y)$. Since
$(X, Y) \in \SE(\Q)$, $Y \models \formGLPb{\P} \circ \formGLPb{\Q}$ and $X \in f(Y)$, we have $(X, Y) \in \SE(\P \star^{\circ, f} \Q)$.
Therefore, $\SE(\P \star \Q) \subseteq \SE(\P \star^{\circ, f} \Q)$. We prove now that $\SE(\P \star^{\circ, f} \Q) \subseteq \SE(\P \star \Q)$.
Let $(X, Y) \in \SE(\P \star \Q)$. We have $(X, Y) \in \SE(\Q)$, $Y \in \formGLPb{\P} \circ \formGLPb{\Q}$ and $X \in f(Y)$. Thus
$(X, Y) \in \SE(\Q)$, $Y \in \min(\modeles{\Q}, \leq_\P)$ and $X \in f(Y)$. We need to show that $X \in \P(Y)$.
Yet since $\P + \Q$ is inconsistent and since we have $(X, Y) \in \SE(\Q)$, we also have $(Y, Y) \in \SE(\Q)$,
so $(Y, Y) \not\in \SE(\P)$, thus $Y \not\models \P$. So by definition of $\P(Y)$, we get that $X \in \P(Y)$.
Since $(X, Y) \in \SE(\Q)$, $Y \in \min(\modeles{\Q}, \leq_\P)$ and $X \in \P(Y)$, we have $(X, Y) \in \SE(\P \star \Q)$.
Therefore, $\SE(\P \star^{\circ, f} \Q) \subseteq \SE(\P \star \Q)$. Hence, $\SE(\P \star^{\circ, f} \Q) = \SE(\P \star \Q)$.
\end{proof}

\noindent {\it Proposition \ref{prop:prop-GLP unique}}~\\
For all propositional-based GLP revision operators $\star^{\circ_1, f_1}, \star^{\circ_2, f_2}$,
we have $\star^{\circ_1, f_1} = \star^{\circ_2, f_2}$ if and only if $\circ_1 = \circ_2$ and $f_1 = f_2$.


\begin{proof}
Let $\star^{\circ_1, f_1}, \star^{\circ_2, f_2}$ be two propositional-based GLP revision operators.\\
\noindent \textit{(If part)} Obvious by Definition \ref{def:propbasedGLPoperator}.\\
\noindent \textit{(Only If part)} Let us prove the contraposite, i.e., assume that $\circ_1 \neq \circ_2$ or $f_1 \neq f_2$
and let us show that $\star^{\circ_1, f_1} \neq \star^{\circ_2, f_2}$.
First, assume that $\circ_1 \neq \circ_2$. This means that there exist two propositional formulae $\phi, \psi$ such that
$\phi \circ_1 \psi \not\equiv \phi \circ_2 \psi$. Then, let $\P, \Q$ be two GLPs defined such that $\formGLPb{\P} \equiv \phi$
and $\formGLPb{\Q} \equiv \psi$.
We have $\modeles{\formGLPb{\P} \circ_1 \formGLPb{\Q}} \neq \modeles{\formGLPb{\P} \circ_2 \formGLPb{\Q}}$.
By Definition \ref{def:propbasedGLPoperator} since $\star^{\circ_1, f_1}, \star^{\circ_2, f_2}$ are
both propositional-based GLP revision operators, $\circ_1$ and $\circ_2$ are both KM revision operators. This means that
$\circ_1$ and $\circ_2$ satisfy the postulate (R2) (see Definition \ref{def:KM revision operator}), but since
$\modeles{\formGLPb{\P} \circ_1 \formGLPb{\Q}} \neq \modeles{\formGLPb{\P} \circ_2 \formGLPb{\Q}}$, this also means
that $\formGLPb{\P} \wedge \formGLPb{\Q}$ is inconsistent, i.e., $\P + \Q$ is inconsistent.
Hence, from Definition \ref{def:propbasedGLPoperator} we can see that for every propositional-based
LP revision operator $\star^{\circ, f}$, we have $\modeles{\P \star^{\circ, f} \Q} = \modeles{\formGLPb{\P} \circ \formGLPb{\Q}}$,
This means that $\modeles{\P \star^{\circ_1, f_1} \Q} \neq \modeles{\P \star^{\circ_2, f_2} \Q}$, thus
$\SE(\P \star^{\circ_1, f_1} \Q) \neq \SE(\P \star^{\circ_2, f_2} \Q)$. Therefore, $\star^{\circ_1, f_1} \neq \star^{\circ_2, f_2}$.\\
Now, assume that $f_1 \neq f_2$. So there exists an interpretation $Y$ such that $f_1(Y) \neq f_2(Y)$.
We fall into at least one of the two following cases: (i) there exists $X \in f_1(Y)$ such that $X \notin f_2(Y)$, or (ii)
there exists $X \in f_2(Y)$ such that $X \notin f_1(Y)$. Assume that we fall into the first case (i) (the second case (ii) leads to
the same result by symmetry). Now, let $\P, \Q$ be two GLPs defined such that $Y \not\models \P$ and $\SE(\Q) = \{(X, Y), (Y, Y)\}$.
$\P + \Q$ is inconsistent. Then by Definition \ref{def:propbasedGLPoperator} we get that
$\SE(\P \star^{\circ_1, f_1} \Q) = \{(X, Y), (Y, Y)\}$ and $\SE(\P \star^{\circ_2, f_2} \Q) = \{(Y, Y)\}$, thus
$\SE(\P \star^{\circ_1, f_1} \Q) \neq \SE(\P \star^{\circ_2, f_2} \Q)$. Therefore, $\star^{\circ_1, f_1} \neq \star^{\circ_2, f_2}$.
\end{proof}

%
%

\noindent {\it Proposition \ref{prop:sigma-injective}}~\\
For every $(\Phi, \Psi) \in GLP_{part}$ and every $\Gamma \in GLP_{faith}$, $((\Phi, \Psi), \Gamma) \in \sigma_{part\rightarrow faith}$
if and only if for all GLPs $\P, \Q$,
$\min(\SE(\Q), \leq^*_\P) = \{(X, Y) \mid (X, Y) \in \SE(\Q),
Y \in \min(\modeles{\Q},$ $\leq_\P),
X \in \P(Y)\}$.

\vspace{2mm}

\begin{proof}
In this proof, for every well-defined set of SE interpretations $S$, $\formwithoutbraces{S}$
denotes any GLP $\P$ such that $\SE(\P) = S$.
Let $(\Phi, \Psi) \in GLP_{part}$ and $\Gamma \in GLP_{faith}$.
We have to show that $((\Phi, \Psi), \Gamma) \in \sigma$, i.e., conditions (i) and (ii) involved in the definition of $\sigma_{part\rightarrow faith}$ are satisfied,
if and only if for all GLP $\P, \Q$, we have $\min(\SE(\Q), \leq^*_\P)$ $= \{(X, Y) \mid (X, Y) \in \SE(\Q),
Y \in \min(\modeles{\Q}, \leq_\P), X \in \P(Y)\}$. For simplicity reasons we abuse notations and respectively denote
$S_{faith} = \min(\SE(\Q), \leq^*_\P)$ and $S_{part} = \{(X, Y) \mid (X, Y) \in \SE(\Q),
Y \in \min(\modeles{\Q}, \leq_\P), X \in \P(Y)\}$.\\

\noindent \textit{(If part)} Assume that for all GLP $\P, \Q$, $S_{faith} = S_{part}$.
We have to show that conditions (i) and (ii) involved in the definition of $\sigma_{part\rightarrow faith}$ are satisfied.

We first prove that (i) for every GLP $\P$ and all interpretations $Y, Y' \in \Omega$, $(Y, Y) \leq^*_\P (Y', Y')$ if and only if $Y \leq_\P Y'$.
Let $Y, Y' \in \allinterpretations$, assume that $(Y, Y) \leq^*_\P (Y', Y')$ and assume toward a contradiction
that $Y' <_\P Y$. Let $\Q$ be the GLP $\Q = \formYYp$. Then $Y \notin \min(mod(\Q), \leq_\P)$, thus $(Y, Y) \notin S_{part}$.
Hence, $(Y, Y) \notin S_{faith}$, which contradicts $(Y, Y) \leq^*_\P (Y', Y')$. The other way around, assume that $Y \leq_\P Y'$ and
assume toward a contradiction that $(Y', Y') <^*_\P (Y, Y)$. Let $\Q$ be the GLP $\Q = \formYYp$. Then $Y \notin S_{faith}$,
thus $Y \notin S_{part}$, which means that $Y \notin \min(mod(\Q), \leq_\P)$ or $Y \notin \P(Y)$. Yet the fact that
$Y \notin \min(mod(\Q), \leq_\P)$ contradicts $Y \leq_\P Y'$ and $Y \notin \P(Y)$ contradicts condition (a) required by the well-defined
assignment $\Psi$. This proves (i).

We now prove that (ii) for every GLP $\P$, $(X, Y) \leq^*_\P (Y, Y)$ and all interpretations $X, Y \in \Omega$ s.t. $X \subseteq Y$,
$(X, Y) \leq^*_\P (Y, Y)$ if and only if $X \in \P(Y)$. Let $X, Y \in \Omega$, $X \subseteq Y$, assume that $(X, Y) \leq^*_\P (Y, Y)$
and assume toward a contradiction that $X \notin \P(Y)$. Then for the GLP $\Q$ defined as $\Q = \formXYY$, we have $(X, Y) \notin S_{faith}$,
so $(X, Y) \notin S_{part}$, which contradicts $(X, Y) \leq^*_\P (Y, Y)$.
The other way around, assume that $X \in \P(Y)$ and assume toward a contradiction that $(Y, Y) <^*_\P (X, Y)$.
Let $\Q$ be the GLP defined as $\Q = \formXYY$. On the one hand $\Q$ has the only model $Y$, so
$\min(mod(\Q), \leq_\P) = \{Y\}$.
On the other hand, we have $(X, Y) \notin S_{faith}$, so $(X, Y) \notin S_{part}$, which means
that we should have $Y \notin \min(mod(\Q), \leq_\P)$ since we assumed that $X \in \P(Y)$. This leads to a contradiction. This proves (ii).\\

\noindent \textit{(Only If part)} Assume that conditions (i) and (ii) involved in the definition of $\sigma_{part\rightarrow faith}$ are satisfied.
We have to show that for all GLP $\P, \Q$, we have $S_{faith} = S_{part}$. Let $\P, \Q$ be two GLPs.

We first prove that $S_{faith} \subseteq S_{part}$. Let $(X, Y) \in S_{faith}$. This means that for every $(X', Y') \in \SE(\Q)$, $(X, Y) \leq^*_\P (X', Y')$.
In particular, $(X, Y) \leq^*_\P (Y', Y')$. And condition (4) required by the GLP compliant faithful assignment $\Gamma$ states that
$(Y, Y) \leq^*_\P (X, Y)$. Hence, $(Y, Y) \leq^*_\P (Y', Y')$. So by condition (i) involved in the definition of $\sigma_{part\rightarrow faith}$,
we get that $Y \leq_\P Y'$ for every $Y' \in \Omega$.
So we showed that $Y \in \min(mod(\Q), \leq_\P)$. Furthermore, since for all $(X', Y') \in \SE(\Q)$, $(X, Y) \leq^*_\P (X', Y')$, we also have that $(X, Y) \leq^*_\P (Y, Y)$,
and condition (ii) involved in the definition of $\sigma_{part\rightarrow faith}$ implies that $X \in \P(Y)$. Since $Y \in \min(mod(\Q), \leq_\P)$ and $X \in \P(Y)$, we get that
$(X, Y) \in S_{part}$.

We prove now that $S_{part} \subseteq S_{faith}$. Let $(X, Y) \in S_{part}$. Since $Y \in \min(mod(\Q)$, $\leq_\P)$, condition (i)
involved in the definition of $\sigma_{part\rightarrow faith}$ implies that $(Y, Y) \leq^*_\P (Y', Y')$ for every $Y' \in \Omega$. Together with condition (4)
required by the GLP compliant faithful assignment $\Gamma$, we get for all $X', Y' \in \Omega$ s.t. $X' \subseteq Y'$ that $(Y, Y) \leq^*_\P (X', Y')$.
And since $X \in \P(Y)$, condition (ii) involved in the definition of $\sigma_{part\rightarrow faith}$ implies that $(X, Y) \leq^*_\P (Y, Y)$. Therefore,
for all $X', Y' \in \Omega$ s.t. $X' \subseteq Y'$, $(X, Y) \leq^*_\P (X', Y')$. This is true in particular for every $(X', Y') \in \SE(\Q)$.
This means that $(X, Y) \in S_{faith}$, and this concludes the proof.
\end{proof}

%
%

\noindent {\it Proposition \ref{prop:ordering-GLP}}~\\
Let $\circ$ be a KM revision operator.
Then for all GLP revision operators $\star_1, \star_2 \in \GLPops{\circ}$,
$\star_1 \lattice{\circ} \star_2$ if and only if for all GLPs $\P, \Q$,
we have $AS(\P \star_1 \Q) \subseteq AS(\P \star_2 \Q)$.

\vspace{2mm}

\begin{proof}
Let $\circ$ be a KM revision operator and $\star_1, \star_2 \in \GLPops{\circ}$.\\
\noindent \textit{(Only if part)} Assume that $\star_1 \lattice{\circ} \star_2$.
By Definition \ref{def:relation GLPs}, for every
interpretation $Y$ we have $f_2(Y) \subseteq f_1(Y)$. Let $\P, \Q$ be two GLPs such that
$\P + \Q$ is inconsistent (the case where $\P + \Q$ is consistent is trivial since by Definition \ref{def:propbasedGLPoperator},
we would have $\P \star_1 \Q = \P \star_2 \Q = \P + \Q$)
and let $Y \in AS(\P \star_1 \Q)$. We need to show that $Y \in AS(\P \star_2 \Q)$. We have
$(Y, Y) \in \SE(\P \star_1 \Q)$ and for every $X \subsetneq Y$, $(X, Y) \notin \SE(\P \star_1 \Q)$.
Since $\star_1$ is a propositional-based revision operator (cf. Proposition \ref{prop:prop-based equl GLP}),
from Definition \ref{def:propbasedGLPoperator} we get that $Y \models \formGLPb{\P} \circ \formGLPb{\Q}$ (i)
and for every $X \subsetneq Y$, $(X, Y) \notin \SE(\Q)$ or $X \notin f_1(Y)$, thus $(X, Y) \notin \SE(\Q)$ or $X \notin f_2(Y)$,
therefore $(X, Y) \notin \SE(\P \star_2 \Q)$ (ii). By (i) we get that $(Y, Y) \in \SE(\P \star_2 \Q)$ and
by (ii) we have for every $X \subsetneq Y$, $(X, Y) \notin SE(\P \star_2 \Q)$. Therefore, by Definition \ref{def:propbasedGLPoperator}
we get that $Y \in AS(\P \star_2 \Q)$. Hence, $AS(\P \star_1 \Q) \subseteq AS(\P \star_2 \Q)$.\\

\noindent \textit{(If part)} Assume that for all GLPs $\P, \Q$, $AS(\P \star_1 \Q) \subseteq AS(\P \star_2 \Q)$.
Toward a contradiction, assume that $\star_1 \not\lattice{\circ} \star_2$. This means that
there exists an interpretation $Y$ such that $f_2(Y) \not\subseteq f_1(Y)$, that is, there exists an interpretation
$X \subsetneq Y$ such that $X \in f_2(Y)$ and $X \notin f_1(Y)$. Then, consider a GLP $\Q$ such that
$\SE(\Q) = \{(X, Y), (Y, Y)\}$ and any GLP $\P$ such that $Y \not\models \P$. Since $Y$ is the only interpretation
satisfying $Y \models \Q$, from postulates (R1) and (R3) of a KM revision operator we have $Y \models \formGLPb{\P} \circ \formGLPb{\Q}$.
Moreover $X \notin f_1(Y)$. So we get from Definition \ref{def:propbasedGLPoperator} that $\SE(\P \star_1 \Q) = \{(Y, Y)\}$.
On the other hand, since $X \in f_2(Y)$ we get that $\SE(\P \star_2 \Q) = \{(X, Y), (Y, Y)\}$. Therefore, $Y \in AS(\P \star_1 \Q)$ and
$Y \notin AS(\P \star_2 \Q)$. This contradicts $AS(\P \star_1 \Q) \subseteq AS(\P \star_2 \Q)$.
\end{proof}

\newpage

\noindent {\it Proposition \ref{prop:sceptical}}~\\
The skeptical GLP revision operators are the only GLP revision operators $\star$ such that
for all GLPs $\P, \Q$,
whenever $\P + \Q$ is inconsistent, we have $AS(\P \star \Q) \subseteq AS(\Q)$.

\vspace{2mm}

\begin{proof}
Let $\circ$ be a KM revision operator and $\star_S^\circ$ be the corresponding skeptical GLP revision operator.
We first show that for all GLPs $\P, \Q$ such that $\P + \Q$ is inconsistent, we have $AS(\P \star_S^\circ \Q) \subseteq AS(Q)$.
$\star_S^\circ$ corresponds to the propositional-based revision GLP operator $\star^{\circ, f}$ such that for every
interpretation $Y$, $f(Y) = 2^Y$. Let $\P, \Q$ be two GLPs such that $\P + \Q$ is inconsistent. Let $Y \in AS(\P \star_S^\circ \Q)$.
We have $(Y, Y) \in \SE(\P \star_S^\circ \Q)$, so by Definition
\ref{def:propbasedGLPoperator} we get that $(Y, Y) \in \SE(\Q)$. Now, assume toward a contradiction that $Y \notin AS(\Q)$.
This means that there exists $X \subsetneq Y$ such that $(X, Y) \in \SE(\Q)$. Yet $f(Y) = 2^Y$, so $X \in f(Y)$, thus by Definition
\ref{def:propbasedGLPoperator} this implies that $(X, Y) \in \SE(\P \star_S^\circ \Q)$, this contradicts $Y \in AS(\P \star_S^\circ \Q)$.
Therefore, $Y \in AS(\Q)$. Hence, $AS(\P \star_S^\circ \Q) \subseteq AS(Q)$.\\
We now show that for some any revision operator $\star^{\circ, f}$, if we have $AS(\P \star \Q) \subseteq AS(Q)$
for all GLPs $\P, \Q$ such that $\P + \Q$ is inconsistent, then $\star^{\circ, f}$ corresponds to the skeptical GLP revision operator
$\star_S^\circ$. Let us show the contraposite, that is, assume that $\star^{\circ, f}$ is not a skeptical GLP revision operator.
This means that there exists an interpretation $Y$ such that $f(Y) \neq 2^Y$, i.e., there exists $X \subsetneq Y$ such that
$X \notin f(Y)$. Then, consider a GLP $\Q$ such that
$\SE(\Q) = \{(X, Y), (Y, Y)\}$ and any GLP $\P$ such that $Y \not\models \P$. Since $Y$ is the only interpretation
satisfying $Y \models \Q$, from postulates (R1) and (R3) of a KM revision operator we have $Y \models \formGLPb{\P} \circ \formGLPb{\Q}$.
On the one hand, since $\SE(\Q) = \{(X, Y), (Y, Y)\}$ we have $Y \notin AS(\Q)$. On the other hand, since $X \notin f(Y)$ we get from Definition
\ref{def:propbasedGLPoperator} that $\SE(\P \star^{\circ, f} \Q) = \{(Y, Y)\}$, that is, $Y \in AS(\P \star^{\circ, f} \Q)$. Therefore,
$AS(\P \star^{\circ, f} \Q) \not\subseteq AS(\Q)$.
\end{proof}

\noindent {\it Proposition \ref{prop:brave}}~\\
The brave GLP revision operators are the only GLP revision operators $\star^{\circ, f}$ such that
for all GLPs $\P, \Q$, whenever $\P + \Q$ is inconsistent, we have
$AS(\P \star^{\circ, f} \Q) = mod(\formGLPb{\P} \circ \formGLPb{\Q})$.

\vspace{2mm}

\begin{proof}
Let $\circ$ be a KM revision operator and $\star_B^\circ$ be the corresponding brave GLP revision operator.
We first show that for all GLPs $\P, \Q$ such that $\P + \Q$ is inconsistent, we have
$AS(\P \star_B^\circ \Q) = mod(\formGLPb{\P} \circ \formGLPb{\Q})$.
$\star_B^\circ$ corresponds to the propositional-based revision GLP operator $\star^{\circ, f}$ such that for every
interpretation $Y$, $f(Y) = \{Y\}$. Let $\P, \Q$ be two GLPs such that $\P + \Q$ is inconsistent. For every interpretation $Y$
and every $X \subsetneq Y$, $X \notin f(Y)$, thus from Definition \ref{def:propbasedGLPoperator} for every interpretation $Y$, we have
$Y \in AS(\P \star_B^\circ \Q)$ if and only if $(Y, Y) \in \SE(\P \star_B^\circ \Q)$ if and only if
$Y \models \formGLPb{\P} \circ \formGLPb{\Q}$. Therefore, $AS(\P \star_B^\circ \Q) = mod(\formGLPb{\P} \circ \formGLPb{\Q})$.\\
We now show that for some any revision operator $\star^{\circ, f}$, if we have
$AS(\P \star_B^\circ \Q) = mod(\formGLPb{\P} \circ \formGLPb{\Q})$
for all GLPs $\P, \Q$ such that $\P + \Q$ is inconsistent, then $\star^{\circ, f}$ corresponds to the brave GLP revision operator
$\star_B^\circ$.
Let us show the contraposite, that is, assume that $\star^{\circ, f}$ is not a brave GLP revision operator.
This means that there exists an interpretation $Y$ such that $f(Y) \neq \{Y\}$, i.e., there exists $X \subsetneq Y$ such that
$X \in f(Y)$. Then, consider a GLP $\Q$ such that
$\SE(\Q) = \{(X, Y), (Y, Y)\}$ and any GLP $\P$ such that $Y \not\models \P$. On the one hand, since $Y$ is the only interpretation
satisfying $Y \models \Q$, from postulates (R1) and (R3) of a KM revision operator we have $Y \models \formGLPb{\P} \circ \formGLPb{\Q}$.
On the other hand, since $\SE(\Q) = \{(X, Y), (Y, Y)\}$ and $X \in f(Y)$, we get from Definition
\ref{def:propbasedGLPoperator} that $\SE(\P \star^{\circ, f} \Q) = \{(X, Y), (Y, Y)\}$, that is, $Y \notin AS(\P \star^{\circ, f} \Q)$. Therefore,
$AS(\P \star_B^\circ \Q) = mod(\formGLPb{\P} \circ \formGLPb{\Q})$.
\end{proof}

\noindent {\it Proposition \ref{prop:drastic coNP}}~\\
$\mathsf{MC}(\circ_D)$ is co$\mathsf{NP}$-complete.

\vspace{2mm}

\begin{proof}
Let $\phi, \psi$ be two formulae and $I$ be an interpretation. In the case where $I \models \phi \wedge \psi$
or $I \not\models \psi$, to determine whether $I \models \phi \circ_D \psi$ can be checked in polynomial time (the answer
is ``yes'' in the former case, ``no'' in the latter one). So let us assume that $I \models \neg\phi \wedge \psi$. Then
to determine whether $I \models \phi \circ_D \psi$ comes down to determine whether $\phi \wedge \psi$ is an inconsistent
formula, that can be down using one call to a co$\mathsf{NP}$ oracle. Hence, $\mathsf{MC}(\circ_D) \in $ co$\mathsf{NP}$.
We prove co$\mathsf{NP}$-hardness by exhibiting a polynomial reduction from the unsatisfiability problem. Consider a propositional formula
$\alpha$ over a set of propositional variables $\mathcal{A}$, and let us associate with it in polynomial time:\\
\noindent $\bullet$ the formulae $\phi, \psi$ defined on $\mathcal{A} \cup \{new, new'\}$ (with $\mathcal{A} \cap \{new, new'\} = \emptyset$) as
$\phi = \alpha \wedge new$ and $\psi = new'$;\\
\noindent $\bullet$ the interpretation $I$ over $\mathcal{A} \cup \{new, new'\}$ defined as $I(p) = 0$ if $p = new$, otherwise $I(p) = 1$.\\
If $\alpha$ is inconsistent then $\phi$ is inconsistent, so $\phi \circ_D \psi = \psi = new'$; since $I(new') = 1$, we get that
$I \models \psi$, so $I \models \phi \circ_D \psi$. Now, if $\alpha$ is consistent then $\phi$ is consistent, so
$\phi \circ_D \psi = \phi \wedge \psi = \alpha \wedge new \wedge new'$; since $I(new) = 0$, we get that $I \not\models \phi \circ_D \psi$.
We just showed that $\alpha$ is inconsistent if and only if $I \models \phi \circ_D \psi$, thus
$\mathsf{MC}(\circ_D)$ is co$\mathsf{NP}$-hard.
\end{proof}

%
%

\noindent {\it Proposition \ref{prop:skeptical-DLP-NLP}}~\\
The skeptical GLP revision operators are both DLP revision operators and NLP revision operators.

\vspace{2mm}

\begin{proof}
We show that every skeptical GLP revision operator $\star^{\circ, f} = \star^{\circ}_S$ is a DLP revision operator.
We have to prove that for all DLP $\P, \Q$, $\P \star^{\circ, f} \Q$ is a DLP, i.e., that
$\SE(\P \star^{\circ}_S \Q)$ is a complete set of SE interpretations. This is trivial when $\P + \Q$ is consistent since in this case,
$\P \star^{\circ}_S \Q = \P + \Q$ and expansion preserves
completeness of SE models, so assume that $\P + \Q$ is inconsistent.
Let $X, Y, Z$ s.t. $Y \subseteq Z$,
$(X, Y), (Z, Z) \in \SE(\P \star^{\circ, f} \Q)$, and let us show that $(X, Z) \in \SE(\P \star^{\circ, f} \Q)$.
By definition of a propositional-based LP revision operator, we know that $(X, Y), (Z, Z) \in \SE(\Q)$. Yet $\Q$ is a DLP, thus
$(X, Z) \in \SE(\Q)$. Since $(Z, Z) \in \SE(\P \star^{\circ, f} \Q)$, we get that $Z \models \formGLPb{\P} \circ \formGLPb{\Q}$.
Moreover, $X \in f(Z)$ since $\star^{\circ}_S$ is a skeptical GLP revision operator. Hence, by definition of a
propositional-based LP revision operator we get that $(X, Z) \in \SE(\P \star^{\circ, f} \Q)$.

One can prove that every skeptical GLP revision operator $\star^{\circ, f} = \star^{\circ}_S$ is a NLP revision operator is a similar way,
by augmenting the above conditions of com-\linebreak pleteness on SE interpretations with the condition of
closeness under here-\linebreak intersection.
\end{proof}

\noindent {\it Proposition \ref{prop:last-result}}~\\
An LP operator $\star$ is a DLP (resp. NLP) revision operator
if and only if there exists a DLP (resp. NLP) parted assignment $(\Phi, \Psi_\Phi)$, where
$\Phi$ associates with every DLP (resp. NLP) $\P$ a total preorder $\leq_\P$,
$\Psi_\Phi$ is a $\Phi$-based complete (resp. normal) assignment
which associates with every DLP (resp. NLP) $\P$ and every interpretation $Y$ a set of interpretations $\P_\Phi(Y)$, and
such that for all DLPs
(resp. NLPs)
$\P, \Q$,
$$\SE(\P \star \Q) = \{(X, Y) \mid (X, Y) \in \SE(\Q),
Y \in \min(\modeles{\Q}, \leq_\P),
X \in \P_\Phi(Y)\}.$$

\vspace{2mm}

\begin{proof}
Let us first prove the representation of DLP revision operators.\\
\textit{(Only if part)} The proof is identical to the one of Proposition \ref{prop:characterization-revision-GLP}
(i.e., our representation theorem for GLP revision operators),
except that we now consider that
for every well-defined set of SE interpretations $S$, $\formwithoutbraces{S}$ denotes any DLP $\R$ whose set of SE models
is the smallest (w.r.t. the set inclusion) superset of $S$, i.e.,
$S \subseteq \SE(\R)$ and there is no DLP $\R'$ such that $S \subseteq \SE(\R')$ and $\SE(\R') \subsetneq \SE(\R)$.
Remark here that given some set $S$, the DLP $\formwithoutbraces{S}$ is uniquely defined (modulo strong equivalence):
to determine $\SE(\formwithoutbraces{S})$, it is enough to add to $S$ all SE interpretations
$(X, Z)$ which are missing from $S$ to ensure its completeness, i.e., those SE interpretations $(X, Z)$ such that $(X, Y), (Z, Z) \in S$ for some interpretation $Y \subseteq Z$.
Also when $S$ is of the form $\{(Y, Y) \mid Y \in E\}$ for some set of interpretations $E$,
we write $\formwithoutbraces{E}$ instead of $\formwithoutbraces{S}$.

Obviously enough, Remark \ref{rem:rem3} and \ref{lemmaRA1RA3GLP} from the proof of Proposition \ref{prop:characterization-revision-GLP}
still hold. We show now that Remark \ref{rem:rem2} from the proof of Proposition \ref{prop:characterization-revision-GLP} also holds, i.e.,
that for all sets of interpretations $E, F$, $\formwithoutbraces{E} + \formwithoutbraces{F} \equiv_s \formwithoutbraces{E \cap F}$.
First, let us show the following intermediate result, that is, for every set $E$ of interpretations and every SE interpretation $(X, Z)$,
\begin{equation}
(X, Z) \in \SE(\formwithoutbraces{E}) \mbox{ if and only if } (X, X), (Z, Z) \in \SE(\formwithoutbraces{E}).
\label{eq:1}
\end{equation}
Equation \ref{eq:1} trivially holds when $X = Z$, so assume $X \subsetneq Z$. The if part comes from the fact
that $\SE(\formwithoutbraces{E})$ is complete. Let us prove the only if part.
On the one hand, $(Z, Z) \in \SE(\formwithoutbraces{E})$ since $\SE(\formwithoutbraces{E})$ is well-defined.
On the other hand, $\SE(\formwithoutbraces{E})$ is complete and minimal w.r.t. the set inclusion, which means that
there necessarily exists $Y \subsetneq Z$, $X \subseteq Y$ such that $(X, Y) \in \SE(\formwithoutbraces{E})$.
If now $X \subsetneq Y$, then the reasoning can be repeated recursively (by setting $Z = Y$ each time). Then after a finite number of
steps we get that $X = Y$ since we deal with a finite set of atoms, that is, $(X, X) \in \SE(\formwithoutbraces{E})$ which proves
that Equation \ref{eq:1} holds. Now, for every SE interpretation $(X, Z)$, we have that 
$$\begin{array}{l}
(X, Z) \in \SE(\formwithoutbraces{E} + \formwithoutbraces{F})\\
\ \ \ \ \mbox{ if and only if } (X, Z) \in \SE(\formwithoutbraces{E}) \cap \SE(\formwithoutbraces{F})\\
\ \ \ \ \mbox{ if and only if } (X, X), (Z, Z) \in \SE(\formwithoutbraces{E}) \cap \SE(\formwithoutbraces{F}) \mbox{ (by Equation \ref{eq:1})}\\
\ \ \ \ \mbox{ if and only if } X, Z \in E \cap F\\
\ \ \ \ \mbox{ if and only if } (X, X), (Z, Z) \in \SE(\formwithoutbraces{E \cap F})\\
\ \ \ \ \mbox{ if and only if } (X, Z) \in \SE(\formwithoutbraces{E \cap F}) \mbox{ (by Equation \ref{eq:1}).}
\end{array}$$
This shows that Remark \ref{rem:rem2} from the proof of Proposition \ref{prop:characterization-revision-GLP} also holds here, i.e.,
that for all sets of interpretations $E, F$, $\formwithoutbraces{E} + \formwithoutbraces{F} \equiv_s \formwithoutbraces{E \cap F}$.

Consider now a DLP revision operator $\revision$.
We associate with $\revision$ a DLP parted assignment $(\Phi, \Psi_\Phi)$ which uses the same construction as for a GLP parted assignment in the
proof of Proposition \ref{prop:characterization-revision-GLP}:
define for every DLP $\P$ the relation $\leq_\P$ over interpretations such that $\forall Y, Y' \in \allinterpretations$, $Y \leq_\P Y'$
iff $Y \models \P \revision \formYYp$, and by defining for every DLP $\P$ and every
$Y \in \allinterpretations$ the set $\P_\Phi(Y)$ as $\P_\Phi(Y) = \{X \subseteq Y \mid \XY \in \SE(\P \revision \formXYY)\}$.
Then the same proof as for Proposition \ref{prop:characterization-revision-GLP} can be used to show that:
\begin{itemize}
\item[(i)] for every DLP $\P$, $\leq_\P$ is a total preorder;
\item[(ii)] for all DLPs $\P, \Q$, $\SE(\P \revision \Q) = \{(X, Y) \mid (X, Y) \in \SE(\Q), Y \in \min(\modeles{\Q},$ $\leq_\P), X \in \P_\Phi(Y)\}$;
\item[(iii)] conditions (1 - 3) of the faithful assignment $\Phi$ and conditions (a - e) of the $\Phi$-based complete assignment $\Psi_\Phi$ are satisfied.
\end{itemize}
It remains to show that the condition (f) of $\Psi_\Phi$ is satisfied. Let $\P$ be a DLP, $X, Y, Z$ be interpretations such that $Y \subseteq Z$, $Y \simeq_\P Z$
and $X \in \P_\Phi(Y)$. Assume toward a contradiction that $X \notin \P_\Phi(Z)$.
By (ii) we get that $\SE(\P \star \form{(X, Y), (Y, Y),$ $(Z, Z), (X, Z)}) = \{(X, Y), (Y, Y), (Z, Z)\}$, which is not a complete set of SE interpretations
since $(X, Z)$ does not belong to it. This contradicts the fact that $\P \star \form{(X, Y), (Y, Y), (Z, Z), (X, Z)}$
is a DLP, i.e., that $\star$ is a DLP revision operator.\\

\noindent\textit{(If part)} We consider a faithful assignment $\Phi$ that associates with every DLP $\P$ a total preorder $\leq_\P$
and a $\Phi$-based complete assignment $\Psi_\Phi$ that associates with every DLP $\P$ and every interpretation
$Y$ a set $\P_\Phi(Y) \subseteq \allinterpretations$. For all DLPs $\P, \Q$, let $\S(\P, \Q)$ be the set of SE interpretations defined as
$\S(\P, \Q) =\{(X, Y) \mid (X, Y) \in \SE(\Q),
Y \in \min(\modeles{\Q}, \leq_\P),
X \in \P_\Phi(Y)\}$. Let $\P, \Q$ be two GLPs. The proof that
$\S(\P, \Q)$ is well-defined is given in the proof of Proposition \ref{prop:characterization-revision-GLP},
by using condition (a) of the $\Phi$-based complete assignment $\Psi_\Phi$.
We show that $\S(\P, \Q)$ is complete by using condition (f). Let $(X, Y), (Z, Z)$ be two SE interpretations such that
$Y \subseteq Z$ and $(X, Y), (Z, Z) \in \S(\P, \Q)$. By definition of $\S(\P, \Q)$ we get that $Y, Z \in \min(\modeles{\Q}, \leq_\P)$,
which means that $Y \simeq_\P Z$, and we also get that $X \in \P_\Phi(Y)$. Thus condition (f) implies that also
$X \in \P_\Phi(Z)$. Therefore, $(X, Z) \in \S(\P, \Q)$ which means that $\S(\P, \Q)$ is complete.
Then we define an operator $\revision$ associating two
DLPs $\P, \Q$ with a new DLP $\P \revision \Q$ such that for all DLPs $\P, \Q$,
$\SE(\P \revision \Q) = \S(\P, \Q)$. The proof that $\revision$ satisfies postulates (RA1 - RA6) is identical to the one
of Proposition \ref{prop:characterization-revision-GLP}.\\

The proof in the NLP case is very similar to the DLP one and uses the same construction, by adapting the structures accordingly and considering the additional
condition (g) involved in a NLP parted assignment.\\
\textit{(Only if part)} For every well-defined set of SE interpretations $S$, $\formwithoutbraces{S}$ denotes any NLP $\R$
(which is uniquely defined modulo equivalence) whose set of SE models
is the smallest (w.r.t. the set inclusion) superset of $S$.
And when $S$ is of the form $\{(Y, Y) \mid Y \in E\}$ for some set of interpretations $E$,
we write $\formwithoutbraces{E}$ instead of $\formwithoutbraces{S}$.

Remark \ref{rem:rem3} and \ref{lemmaRA1RA3GLP} from the proof of Proposition \ref{prop:characterization-revision-GLP}
still hold, but we need to show that Remark \ref{rem:rem2} from the proof of Proposition \ref{prop:characterization-revision-GLP} also holds, i.e.,
that for all sets of interpretations $E, F$, $\formwithoutbraces{E} + \formwithoutbraces{F} \equiv_s \formwithoutbraces{E \cap F}$.
For this purpose, we prove an adaptation of Equation \ref{eq:1} previously given in this proof for DLPs, to the case of NLPs; that is,
for every set $E$ of interpretations and every SE interpretation $(X, Z)$,
\begin{equation}
\hspace*{-1cm}\begin{array}{l}
(X, Z) \in \SE(\formwithoutbraces{E}) \mbox{ if and only if one of the two following conditions holds:}\\
\begin{array}{ll}
(i) & (X, X), (Z, Z) \in \SE(\formwithoutbraces{E})\\
(ii) & \mbox{there is a set of interpretations } \mathcal{Y} \mbox{ such that } \bigcap_{Y \in \mathcal{Y}}{Y} = X, |\mathcal{Y}| \geq 2,\\
& \mbox{and } \forall Y \in \mathcal{Y}, (Y, Y) \in \SE(\formwithoutbraces{E}).
\end{array}
\end{array}
\label{eq:2}
\end{equation}

Equation \ref{eq:2} trivially holds when $X = Z$,
so assume $X \subsetneq Z$. The if part comes from the fact
that $\SE(\formwithoutbraces{E})$ is complete and closed under here-intersection. Let us prove the only if part.
Assume that it does not hold that $(X, X), (Z, Z) \in \SE(\formwithoutbraces{E})$.
Then by Equation \ref{eq:1}, $(X, Z)$ belongs to $\SE(\formwithoutbraces{E})$ because its condition specific for
closure under here-intersection, i.e., $\exists Y, Y' \subseteq Z$, $Y \cap Y' = X$, $Y \neq Y'$, $(Y, Z), (Y', Z) \in \SE(\formwithoutbraces{E})$.
By applying this reasoning recursively, since we are dealing with a finite set of atoms
there must exist a finite set $\mathcal{Y}$ of at least two interpretations such that $\bigcap_{Y \in \mathcal{Y}}{Y} = X$,
and such that all $(Y, Z)$ such that $Y \in \mathcal{Y}$ belong $\SE(\formwithoutbraces{E})$ because the condition of completeness,
which means by Equation \nolinebreak\ref{eq:1} that for every $Y \in \mathcal{Y}$, $(Y, Y) \in \SE(\formwithoutbraces{E})$.

Now, for every SE interpretation $(X, Z)$, we have that 
$$\begin{array}{l}
(X, Z) \in \SE(\formwithoutbraces{E} + \formwithoutbraces{F})\\
\ \ \ \ \mbox{ if and only if } (X, Z) \in \SE(\formwithoutbraces{E}) \cap \SE(\formwithoutbraces{F})\\
\ \ \ \ \mbox{ if and only either (i) or (ii) from Equation \ref{eq:2} holds for both $E$ and $F$}.
\end{array}$$
Yet on the one hand, condition (i) from Equation \ref{eq:2} holds for both $E$ and $F$ if and only if $X, Z \in E \cap F$ if and only if $(X, X), (Z, Z) \in \SE(\formwithoutbraces{E \cap F})$.
On the other hand, condition (ii) from Equation \ref{eq:2} holds for both $E$ and $F$
if and only if
there is a set of interpretations $\mathcal{Y}$ such that $\bigcap_{Y \in \mathcal{Y}}{Y} = X$, $|\mathcal{Y}| \geq 2$
and $\forall Y \in \mathcal{Y}, (Y, Y) \in \SE(\formwithoutbraces{E}) \cap \SE(\formwithoutbraces{F})$,
if and only
there is a set of interpretations $\mathcal{Y}$ such that $\bigcap_{Y \in \mathcal{Y}}{Y} = X$, $|\mathcal{Y}| \geq 2$
and $\forall Y \in \mathcal{Y}$, $Y \in E \cap F$,
if and only if 
there is a set of interpretations $\mathcal{Y}$ such that $\bigcap_{Y \in \mathcal{Y}}{Y} = X$, $|\mathcal{Y}| \geq 2$
and $\forall Y \in \mathcal{Y}$, $(Y, Y) \in \SE(\formwithoutbraces{E \cap F})$.
Therefore, by Equation \ref{eq:2} we get that either (i) or (ii) from Equation \ref{eq:2} holds for both $E$ and $F$
if and only if $(X, Z) \in \SE(\formwithoutbraces{E \cap F})$.
This shows that Remark \ref{rem:rem2} from the proof of Proposition \ref{prop:characterization-revision-GLP} also holds here, i.e.,
that for all sets of interpretations $E, F$, $\formwithoutbraces{E} + \formwithoutbraces{F} \equiv_s \formwithoutbraces{E \cap F}$.

Consider now a NLP revision operator $\revision$, and similarly to the case of DLPs,
we associate with $\revision$ the following NLP parted assignment $(\Phi, \Psi_\Phi)$:
we define for every DLP $\P$ the relation $\leq_\P$ over interpretations such that $\forall Y, Y' \in \allinterpretations$, $Y \leq_\P Y'$
iff $Y \models \P \revision \formYYp$, and for every GLP $\P$ and every
$Y \in \allinterpretations$ the set $\P_\Phi(Y)$ as $\P_\Phi(Y) = \{X \subseteq Y \mid \XY \in \SE(\P \revision \formXYY)\}$.
Then the same proof as for Proposition \ref{prop:characterization-revision-GLP} can be used to show that:
\begin{itemize}
\item[(i)] for every NLP $\P$, $\leq_\P$ is a total preorder;
\item[(ii)] for all NLPs $\P, \Q$, $\SE(\P \revision \Q) = \{(X, Y) \mid (X, Y) \in \SE(\Q), Y \in \min(\modeles{\Q},$ $\leq_\P), X \in \P_\Phi(Y)\}$;
\item[(iii)] conditions (1 - 3) of the faithful assignment $\Phi$ and conditions (a - e) of the $\Phi$-based complete assignment $\Psi_\Phi$ are satisfied.
\end{itemize}
Additionally, we can use the same proof as for DLPs to show that condition (f).
It remains to show that the condition (g) of $\Psi_\Phi$ is satisfied.
Let $\P$ be a DLP, $X, Y, Z$ be interpretations such that
$X, Y \in \P_\Phi(Z)$. Assume toward a contradiction that $X \cap Y \notin \P_\Phi(Z)$.
By (ii) we get that $\SE(\P \star \form{(X, Z), (Y, Z), (Z, Z), (X \cap Y, Z)}) = \{(X, Y), (Y, Y), (Z, Z)\}$, which is not closed under here-intersection
since $(X \cap Y, Z)$ does not belong to it. This contradicts the fact that $\P \star \form{(X, Z), (Y, Z),$ $(Z, Z), (X \cap Y, Z)}$
is a NLP, i.e., that $\star$ is a NLP revision operator.\\

\noindent\textit{(If part)} We consider a NLP parted assignment $(\Phi, \Psi_\Phi)$ defined as the DLP parted assignment in the if part of the proof for the DLP case.
Then defined an operator $\revision$ associating two
NLPs $\P, \Q$ with a new NLP $\P \revision \Q$ such that for all NLPs $\P, \Q$,
$\SE(\P \revision \Q) = \{(X, Y) \mid (X, Y) \in \SE(\Q),
Y \in \min(\modeles{\Q}, \leq_\P),
X \in \P_\Phi(Y)\}$.
We already showed that $\SE(\P \revision \Q)$ is well-defined and complete, and condition (g) of the $\Phi$-based normal assignment $\Psi_\Phi$
directly implies that $\SE(\P \revision \Q)$ is closed under here-intersection. Therefore, $\P \revision \Q$ is a NLP.
The proof that $\revision$ satisfies postulates (RA1 - RA6) is identical to the one
of Proposition \ref{prop:characterization-revision-GLP}.
\end{proof}
\end{document}